\documentclass[10pt,twocolumn,a4paper]{article}

\usepackage{iftex}
\ifPDFTeX
  \usepackage[T1]{fontenc}
  \usepackage[utf8]{inputenc}
  \usepackage{mathpazo}   
\fi
\usepackage[spanish,es-tabla,english]{babel}
\usepackage{csquotes}
\usepackage{microtype}

\usepackage[a4paper,margin=2cm]{geometry}
\usepackage{setspace}
\usepackage{graphicx}
\usepackage{booktabs}
\usepackage{array}
\usepackage{amsmath}
\usepackage{amssymb}
\usepackage{dsfont}      
\usepackage{caption}
\usepackage{subcaption}
\usepackage{float}
\usepackage{enumitem}
\usepackage{hyperref}
\usepackage{isnimeta}
\usepackage{url}
\usepackage{xcolor}
\usepackage{lastpage}

\hypersetup{
  colorlinks=true,
  linkcolor=black,
  citecolor=black,
  urlcolor=black,
  pdfauthor={Nicolás Rodríguez Álvarez},
  pdftitle={Hallucination as a Feature, not a Defect}
}

\isnisetup{
  metadata = true,
  xmp = false,
  export = true,
  export-format = json
}

\isnimetadata{
  title = {Hallucination as a Feature, not a Defect},
  publisher = {arXiv},
  publication-year = {2026},
  resource-type = {Preprint},
  language = {en}
}

\usepackage[
  backend=biber,
  style=authoryear,
  natbib=true,
  maxcitenames=2,
  maxbibnames=99,
  giveninits=true,
  uniquename=false,
  doi=true,
  url=true,
  isbn=false
]{biblatex}
\newcommand{\sourceNote}[1]{\caption*{\footnotesize Source: #1}}

\title{\textbf{Hallucination as a Feature, not a Defect}\\[0.3em]
\large Evaluating a multi-agent architecture to transform speculative
language-model outputs into testable scientific hypotheses}
\author{\isniauthor[
  given={Nicolás},
  family={Rodríguez Álvarez},
  affiliation={IES Parquesol, Valladolid, Castilla y León, Spain},
  orcid={0009-0002-6804-386X},
  role=author,
  type=person
]{0000000530382326}\\
\small IES Parquesol, Valladolid, Castilla y León, Spain\\
\small Technology and Engineering}
\date{2026}

\begin{document}

\twocolumn[
\begin{@twocolumnfalse}
\maketitle
\begin{abstract}
\noindent Contemporary Large Language Models (LLMs) are increasingly aligned to suppress hallucinations, prioritizing factual retrieval over combinatorial creativity. While crucial for mitigating misinformation, this alignment may also restrict speculative Research and Development (R\&D) by encouraging what this work operationally treats as semantic overfitting and diversity collapse. In this paper, we propose a Rust-based multi-agent orchestration that uses the contrast between narrative daydreaming and executive control as a functional analogy, not as a neurocognitive claim. The system instigates an \textit{Epistemological Friction} loop between a high-entropy generating agent and a web-grounded evaluating agent, mediated by a low-entropy semantic bottleneck intended to reduce noise and repetition. Initial experiments generated diverse, viability-rated hypotheses across physical and social-science domains. We additionally report an exploratory paired baseline and ablation study comparing the full system against direct prompting, self-reflection, removal of the semantic filter, removal of search grounding, and removal of lateral lenses. The results place direct prompting among the weakest conditions across most observed metrics, but they do not show a general superiority of the full system over simple self-reflection. Instead, they suggest that each architecture shifts the balance between originality, feasibility, diversity, and empirical grounding in different ways, and that the full system provides its main advantages when hypotheses must survive strong physical, empirical, or institutional constraints. These findings do not show that hallucination is useful in isolation; they suggest that speculative generation gains value only when constrained by architecture, empirical grounding, and explicit evaluation.


\smallskip
\noindent\textbf{Note on language versions.} Following arXiv's multilingual submission guidelines, this document contains the full English version first, followed by the complete Spanish version (\textit{versión en español}).
\end{abstract}
\bigskip
\end{@twocolumnfalse}
]

\section{Introduction}
As a conceptual starting point, this work draws a functional analogy from the literature on intelligence and cognition: the tension between narrative, associative, or contemplative processes and goal-directed control processes \parencite{nilsson_quest_2009}. The reference to the Default Mode Network (DMN) is not used here as a strong neuroscientific hypothesis, but as an operational metaphor to separate two computational functions: broad semantic exploration and evaluative constraint.

To this end, we propose a multi-agent architecture that translates this analogy into a technical design. A generating agent produces high-entropy associations; other modules introduce empirical critique, semantic filtering, and final evaluation. The aim is not to claim that LLMs reproduce human cognition, but to test whether an explicit separation between speculation and control improves the production of structured hypotheses.

\section{Background}
Paradoxically, the recent trajectory of artificial intelligence research attempts, with all its resources, to suppress an inherent capability of current Large Language Models: the propensity to \enquote{hallucinate} \parencite{chen_combating_2024,ji_survey_2023}. The major laboratories perceive this phenomenon fundamentally as a vector that aggravates AI-generated misinformation, producing responses that diverge from truthful information. As a result, substantial efforts have been deployed in alignment, reinforcement learning from human feedback (RLHF), and rigorous empirical grounding to ensure that models act as reliable information retrievers rather than speculative engines \parencite{ji_survey_2023}.

While it is undeniable that mitigating misinformation is critical for public-facing applications \parencite{chen_combating_2024}, this relentless optimization of factual adherence carries a serious cognitive cost: it curtails the model's capacity for divergent, out-of-distribution combinatorial ideation. By treating all hallucinations uniformly as a \enquote{defect} to be eliminated, the discipline inadvertently neutralizes the very mechanism that allows LLMs to make associations across distant semantic domains.

In design terms, a limited analogy can be drawn between high-entropy autoregressive generation and associative thinking: both can produce non-obvious connections, but also errors, sycophancy, or physically infeasible proposals. Conversely, alignment oriented solely toward avoiding risk may favor more conservative responses. The central question is not whether LLMs possess a biological equivalent of the DMN, but whether an architecture that separates speculative generation from critical control can harness part of that divergence without accepting its outputs unverified.

\section{Working hypothesis and research objectives}

\subsection{Working hypothesis}
We propose an architectural hypothesis, not a neurocognitive one: a high-entropy generative phase can increase hypothesis exploration if it is explicitly separated from a subsequent phase of critique, filtering, and evaluation. The prediction is not that hallucination is beneficial in itself, but that controlled speculation can change the profile of the hypotheses produced.

\subsection{Objectives}
In this paper, the main objective is to present a Rust-based multi-agent R\&D orchestrator that implements this functional separation. By forcing \enquote{epistemological friction} between a speculative generating agent and a search-based empirical evaluating agent, we seek to assess whether a divergent generation phase, subsequently subjected to critique and filtering, produces more structured hypotheses than direct prompting of the model.

Unlike standard goal-oriented multi-agent frameworks designed for task execution and tool use (such as ReAct \parencite{yao_react_2023} or AutoGPT \parencite{yang_auto-gpt_2023}), our architecture seeks to be optimized strictly for speculative theoretical friction.

\section{Materials and methods}

\subsection{Overall system design}
The system studies whether the speculative generation of language models can be used as a controlled phase of scientific ideation. The goal is not to directly accept the model's unverified outputs, but to treat them as candidate hypotheses that must subsequently be criticized, filtered, and evaluated.

The architecture was designed as a multi-agent system. Instead of requesting a single direct response from an LLM, the process is divided into distinct functions: hypothesis generation, critique, refinement, semantic filtering, and evaluation. This decision relates to recent work on iterative refinement through the model's own feedback, such as \textit{Self-Refine} and \textit{Reflexion}, and to multi-agent debate approaches aimed at improving reasoning and factuality \parencite{madaan_self-refine_2023,shinn_reflexion_2023,du_improving_2023}.

The full version of the system includes several roles: a module that configures the problem domain, a hypothesis-generating agent, a critic agent, a semantic filter, and a final evaluator. The central idea is to create a tension between exploration and constraint. We call this tension \textit{epistemological friction}: a process in which speculative ideas are subjected to criteria of feasibility, specificity, non-duplication, and novelty before being considered useful results.

\begin{figure*}[!tp]
    \centering
    \includegraphics[width=0.7\textwidth]{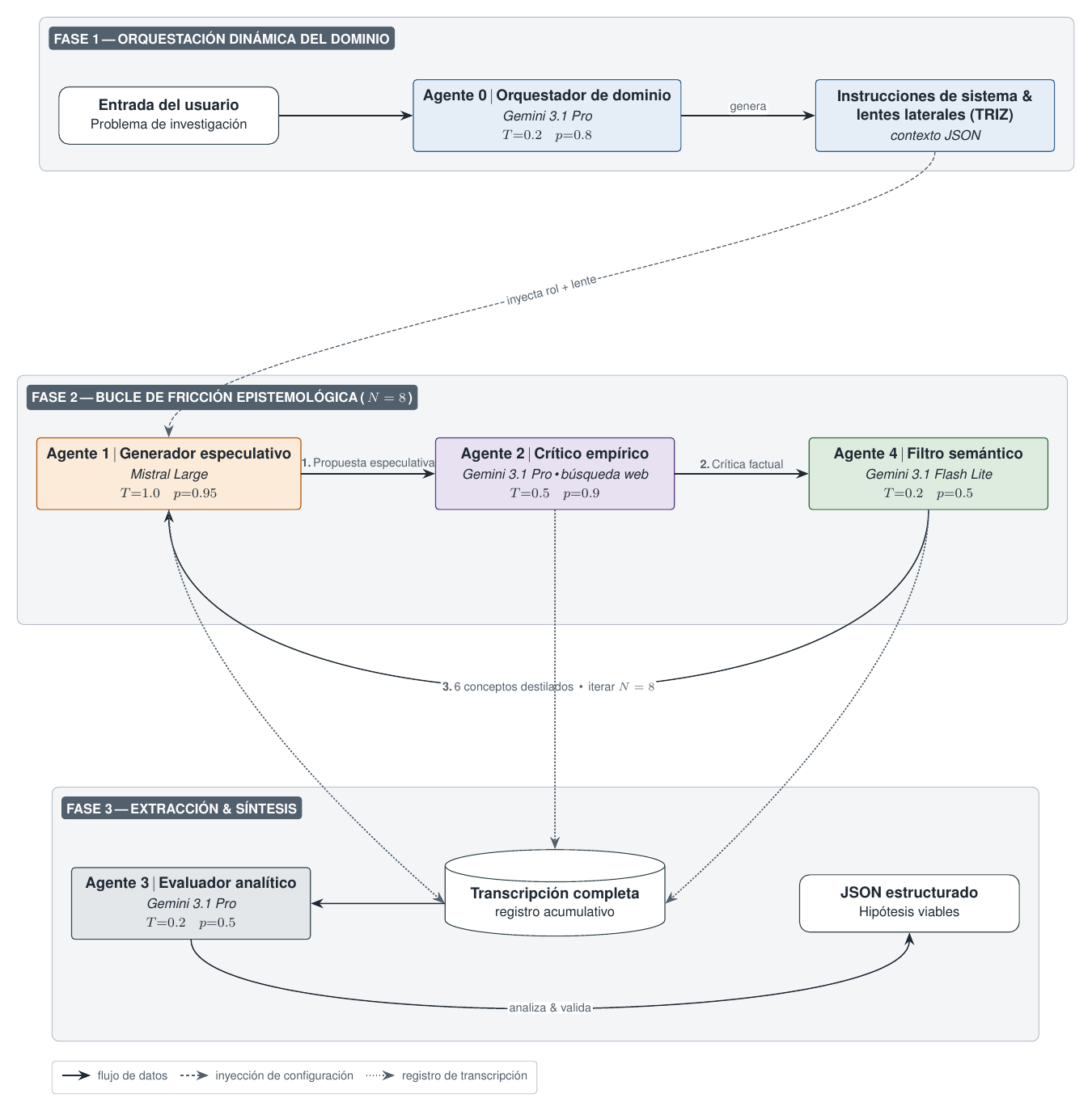}
    \caption{Architecture of the multi-agent pipeline and data flow between configuration, the epistemological-friction loop, and final evaluation.}
    \label{fig:pipeline-en}
    \sourceNote{own elaboration, based on the TikZ diagram and the orchestrator implemented for this study.}
\end{figure*}

\noindent\textbf{Operational scheme.} The complete flow is: (1) problem input and constraints; (2) automatic configuration of the domain, system instructions, and lateral lenses; (3) speculative generation by the creator agent; (4) empirical critique with or without external search, depending on the experimental condition; (5) semantic filtering of the critique and feedback over eight rounds; (6) final extraction of hypotheses in JSON; and (7) reproducible computation of metrics over the experimental logs. Figure~\ref{fig:pipeline-en} summarizes this architecture.

\subsection{Models, tools, and execution environment}
The system was implemented primarily in Rust. The orchestration, ablation conditions, choice of lateral lenses, and log writing are deterministic for a given configuration and seed; the concrete LLM responses remain stochastic because they depend on external inference services. Each run generates structured JSON logs, later transformed into tables for analysis.

The models used in the experimental run were: \path{mistral-large-latest} for the generating agent; \path{gemini-3.1-pro-preview} via Vertex AI for domain configuration, critique, and evaluation; and \path{gemini-3.1-flash-lite-preview} for the semantic filter. The parameters were: generator \(T=1.0\), \(\textit{top-p}=0.95\); critic \(T=0.5\), \(\textit{top-p}=0.9\); evaluator \(T=0.2\), \(\textit{top-p}=0.5\); architect \(T=0.2\), \(\textit{top-p}=0.8\); extractor/filter \(T=0.2\), \(\textit{top-p}=0.5\). Vertex AI web search was enabled only in the conditions with external grounding.

The quantitative analysis was carried out with the Julia package \texttt{HallucinationAsAFeatureJulia} \parencite{rodriguez_alvarez_2026_20649714}. From the logs we generated a run manifest, a table of individual hypotheses, aggregated metrics per run, metrics per experimental condition, paired differences, and average rankings. This separation between execution and analysis allows the metrics to be recomputed without repeating all the generative experiments.

\subsection{Reproducible implementation of the metrics}
Feasibility was not annotated manually. It was assigned by the final evaluator (Agent 3) through a mandatory JSON output with the field \texttt{feasibility\_rank}. The conversion was deterministic: \textit{low}=1, \textit{medium}=2, and \textit{high}=3; numeric values, when present, were clamped to the interval 0--3. Operationally, a score of 1 indicates an idea with dominant technical or institutional barriers; 2, a plausible idea but with significant implementation or validation risks; and 3, an idea with a clear mechanism and a reasonable test path. This decision introduces partial circularity because an LLM evaluates LLM outputs; therefore, feasibility is interpreted as a rubric score and not as independent expert validation. Strict scientific validation would require specialist review, experiments, or comparison with external datasets.

The specificity variables were computed deterministically. The domain is inherited from the experimental problem; the presence of a mechanism, a measurable variable, and testability is detected first through explicit fields if present and, failing that, through fixed lexical patterns. For example, mechanism is inferred from lexical families associated with causal pathway, mechanism, effect, use, or gradient; the measurable variable from magnitudes, units, performance, temperature, pressure, energy, precision, or thresholds; and testability from terms related to test, measurement, experiment, simulation, comparison, observability, dataset, or model.

The distance between hypotheses was not computed with an external embedding model, but with sparse token-count vectors and cosine similarity. Tokens were lowercased and alphanumeric words of three or more characters were retained. Therefore, the metric should be read as a reproducible lexical-semantic proxy, not as a deep measure of scientific novelty.

\subsection{Experimental design}
The study was framed as a comparison between the full system, several baselines, and different ablations. An ablation consists of removing or modifying a component of the system to observe how the overall behavior changes. This makes it possible to estimate the contribution of each module to hypothesis generation.

Six experimental conditions were compared:

\begin{itemize}[leftmargin=*]
    \item \textit{direct\_llm}: direct response from a language model without a multi-agent architecture.
    \item \textit{self\_reflection}: simple generation and revision by the model itself, inspired by iterative refinement methods such as \textit{Self-Refine} and \textit{Reflexion} \parencite{madaan_self-refine_2023,shinn_reflexion_2023}.
    \item \textit{no\_agent4}: system without the semantic filter.
    \item \textit{no\_search}: system without the external empirical search or critique phase.
    \item \textit{no\_lateral\_lenses}: system without alternative lateral perspectives.
    \item \textit{full\_system}: the complete multi-agent architecture.
\end{itemize}

The planned experimental matrix was:

\[
\begin{aligned}
& 3 \text{ problem--seed blocks} \times 6 \text{ conditions} \\
&\quad \times 10 \text{ repetitions} = 180 \text{ expected runs}.
\end{aligned}
\]

The blocks were: desalination with seed 42, desalination with seed 43, and parliamentary deadlock with seed 42. The ten repetitions per condition reduce dependence on a single generative output, but they do not turn the study into 180 independent problems. Moreover, by evaluating different conditions on the same problem--seed blocks, paired comparisons can be performed. The primary unit of descriptive analysis is the complete run; the paired comparisons, when aggregated by problem--seed block, remain exploratory because there are only three effective blocks.

\subsection{Evaluation metrics}
The evaluation was designed as a multidimensional framework. A single creativity score is not used, because a hypothesis can be novel but infeasible, or feasible but trivial. This decision is supported by recent benchmarks such as \textit{LiveIdeaBench}, which evaluates scientific idea generation through dimensions such as originality, feasibility, fluency, flexibility, and clarity, and \textit{CreativityPrism}, which separates LLM creativity into quality, novelty, and diversity \parencite{ruan_evaluating_2026,hou_creativityprism_2026}.

Let \(H_r\) be the set of hypotheses generated in a run \(r\), and let \(i\) be an individual hypothesis. Productivity or fluency is defined as:

\[
Fluency_r = |H_r|.
\]

Feasibility was coded as a normalized ordinal score:

\[
FeasibilityNorm_i = \frac{Feasibility_i}{3}.
\]

The score \(Feasibility_i\) comes from the \texttt{feasibility\_rank} field assigned by Agent 3 and transformed in a fixed way: \(\textit{low}=1\), \(\textit{medium}=2\), \(\textit{high}=3\).

A hypothesis is considered feasible with a score of 2 or higher:

\[
Feasible_i = \mathds{1}[Feasibility_i \geq 2].
\]

Specificity measures whether the hypothesis contains four minimal elements: domain, mechanism, measurable variable, and testability. It is computed as:

\[
\begin{aligned}
Specificity_i = \tfrac{1}{4}\big(
& Domain_i + Mechanism_i \\
&+ MeasurableVariable_i + Testability_i \big).
\end{aligned}
\]

A hypothesis is considered specific when:

\[
Specific_i = \mathds{1}[Specificity_i \geq 0.75].
\]

We also assess whether the hypothesis is logically possible and whether it is non-duplicate. Non-duplication is estimated through semantic similarity among hypotheses of the same run:

\[
NotDuplicate_i = \mathds{1}[MaxSimilarity_i < 0.90].
\]

In this implementation, \(MaxSimilarity_i\) is the maximum cosine similarity between the token-count vector of hypothesis \(i\) and that of any other hypothesis in the same run.

With these criteria, structural validity is defined:

\[
\begin{aligned}
StructuralValid_i = {}
& Specific_i \land Feasible_i \\
&\land LogicalPossible_i \land NotDuplicate_i.
\end{aligned}
\]

Originality was estimated through lexical-semantic distance with respect to the baselines. Let \(v_i\) be the token-count vector of hypothesis \(i\). The distance used in all metrics is:

\[
d(i,j)=1-\cos(v_i,v_j).
\]

For each hypothesis, its distance to the nearest neighbor generated by the baseline conditions is computed:

\[
NBN_i = \min_{j \in B_p} d(i,j),
\]

where \(B_p\) is the set of \textit{direct\_llm} and \textit{self\_reflection} hypotheses associated with the same problem \(p\), excluding the file itself where appropriate.

To avoid a completely arbitrary threshold, high novelty was defined relative to the upper quartile of the novelty distribution of the direct model:

\[
HighNovel_i = \mathds{1}\!\left[ NBN_i \geq Q_{0.75}(NBN_{direct\_llm}) \right].
\]

The main metric of the analysis was the number of structurally valid and highly novel hypotheses per run:

\[
\begin{aligned}
SVHN_r = \sum_{i \in H_r} \mathds{1}\big[\,
& StructuralValid_i \\
&\land HighNovel_i \,\big].
\end{aligned}
\]

This metric measures how many hypotheses in a run are simultaneously specific, feasible, logically possible, non-redundant, and far from the baseline conditions according to the lexical-semantic proxy used.

Combinatorial rarity was also computed:

\[
CombinationRarity_i = -\log(p(t_i)),
\]

where \(t_i\) represents the combination of problem and disciplines associated with the hypothesis. The analysis applies additive smoothing, \(p(t_i)=(count(t_i)+1)/(N+K)\), and then normalizes the result to the interval \([0,1]\) according to the observed minimum and maximum. The least frequent combinations within the experimental set receive larger values.

Semantic diversity was computed as the mean distance between pairs of hypotheses within a run:

\[
SemanticDiversity_r =
\frac{2}{n_r(n_r-1)}
\sum_{i<j} d(i,j).
\]

This metric is only computed when a run contains at least two hypotheses. Finally, semantic collapse was estimated as the proportion occupied by the largest cluster of hypotheses. Clusters were built as connected components: two hypotheses are joined if their cosine similarity is 0.90 or higher.

\[
CollapseRate_r =
\frac{\text{size of the largest cluster}}{n_r}.
\]

High values indicate that many hypotheses concentrate on a single idea; low values indicate greater conceptual dispersion. As with diversity, collapse is only computed for runs containing at least two hypotheses: with a single hypothesis the largest cluster trivially spans the whole run, and the metric would confound low productivity with semantic concentration.

\subsection{Analysis procedure}
The analysis was carried out in four phases. First, a run manifest was created with the problem, seed, condition, output file, and status of each run. This made it possible to compute completion rates and to separate technical failures from semantic results.

Second, individual hypotheses were extracted from the experimental logs and a table was built in which each row corresponds to a hypothesis. On this table, the metrics of feasibility, specificity, novelty, combinatorial rarity, diversity, and non-duplication were computed.

Third, the individual metrics were aggregated by run and then by experimental condition. This yielded mean values per condition for comparing the full system with the baselines and the ablations.

Fourth, paired comparisons were computed. For a metric \(m\), the difference between the full system and a baseline within the same problem--seed block is defined as:

\[
\Delta_{b,m} =
m_{\text{system},b} - m_{\text{baseline},b}.
\]

This analysis is more informative than comparing only global means, because it partially controls for problem difficulty and the variability introduced by the seed. However, it must be interpreted strictly descriptively: although there were 180 runs, the paired differences are aggregated by problem--seed block and only three effective blocks exist. Therefore, the effect sizes, win rates, and confidence intervals in the appendix are included as exploratory indicators of direction and internal stability, not as robust statistical inference or as a test of population significance.

In this version, a complete claim-based grounding verification was not performed for all hypotheses. For this reason, the main metric is called structural validity, not definitive scientific validation.

\section{Results}

\subsection{Quantitative results}
The quantitative results are summarized through two main visualizations. The first shows a composite index of feasible combinatorial creative performance, used as a proxy to compare each condition's capacity to produce novel, feasible, and combinatorially rich hypotheses. The second compares the conditions across several normalized dimensions: structural validity, feasibility, originality, combinatorial rarity, diversity, and semantic collapse.

\begin{figure}[!htbp]
    \centering
    \includegraphics[width=\linewidth]{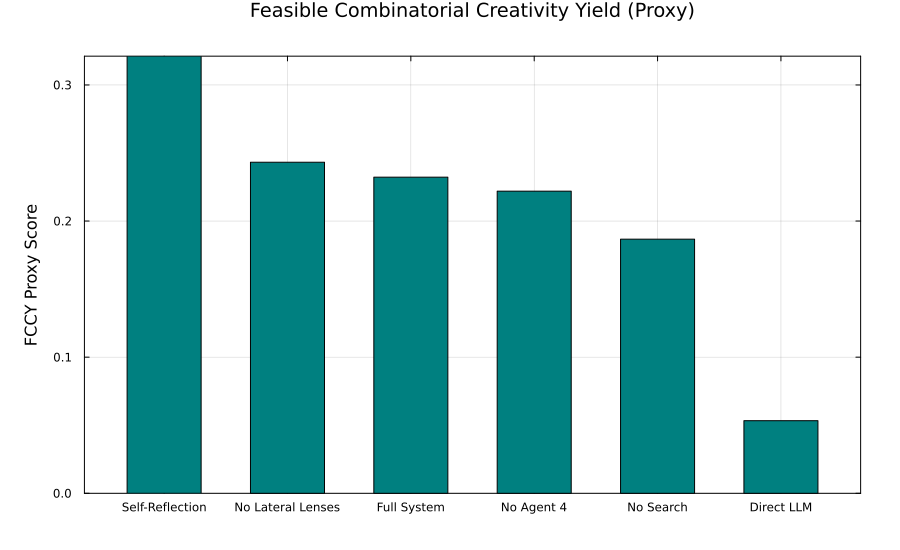}
    \caption{Feasible combinatorial creative performance by experimental condition. The FCCY index is interpreted as a proxy measure of each architecture's capacity to produce hypotheses that combine feasibility, originality, and combinatorial rarity.}
    \label{fig:fccy-en}
    \sourceNote{own elaboration.}
\end{figure}

As shown in Figure~\ref{fig:fccy-en}, the conditions based on reflection or structured architecture clearly outperform the direct model. This result suggests that introducing review, filtering, or structured generation processes can increase the creative performance measured by the composite index. Nonetheless, this index should be interpreted as an auxiliary measure, not as definitive scientific validation of the generated hypotheses.

\begin{figure*}[!tp]
    \centering
    \includegraphics[width=0.95\textwidth]{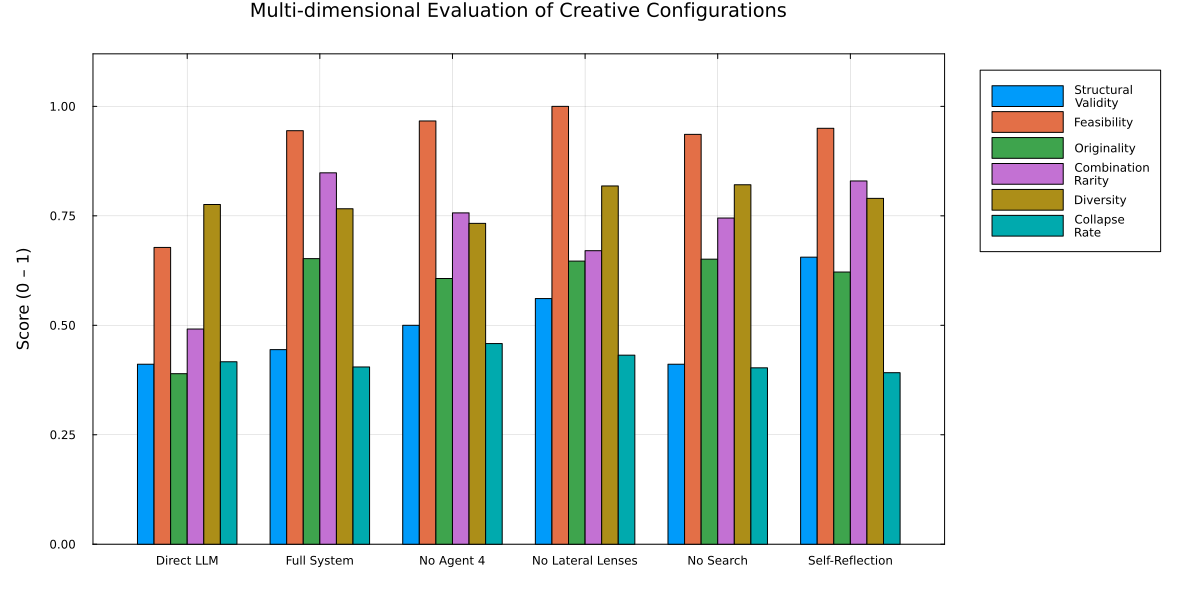}
    \caption{Multidimensional evaluation of the experimental conditions. Six dimensions are compared: structural validity, feasibility, originality, combinatorial rarity, diversity, and semantic collapse rate (the latter computed only over runs with at least two hypotheses).}
    \label{fig:multidimensional-results-en}
    \sourceNote{own elaboration.}
\end{figure*}

Figure~\ref{fig:multidimensional-results-en} shows that no condition absolutely dominates all metrics. The direct model obtains lower results in most dimensions, while the architectures with reflection or structured ablations show improvements in originality, combinatorial rarity, and diversity. This pattern partially supports the initial hypothesis: speculative generation can be useful as an exploration phase, but its value depends on subsequent filters of feasibility, specificity, and non-redundancy.

The complete numerical values corresponding to these figures are collected in the additional tables of the appendix.

\subsection{Qualitative results}
In addition to the aggregated metrics, real outputs from the same experiments were reviewed manually. Table~\ref{tab:qualitative-examples-en} shows representative contrasts between the direct model, simple self-reflection, and the full system. In both problems, the direct model tends to produce a striking but weakly constrained idea; self-reflection improves the form and local plausibility, while the full system introduces a more external critical pressure that shifts the proposal toward more specific or institutionally defensible mechanisms.

\begin{table*}[!tp]
\centering
\scriptsize
\caption{Real qualitative examples of epistemological friction between \textit{direct\_llm}, \textit{self\_reflection}, and \textit{full\_system}.}
\label{tab:qualitative-examples-en}
\resizebox{\textwidth}{!}{%
\begin{tabular}{p{2.4cm}p{2.5cm}p{6.4cm}p{5.8cm}}
\toprule
\textbf{Problem} &
\textbf{Condition} &
\textbf{Generated hypothesis} &
\textbf{Qualitative reading} \\
\midrule
Desalination &
\textit{direct\_llm} &
\textbf{Electromagnetic pumping of seawater in the solid state.} Proposes using crossed electric and magnetic fields (MHD) or traveling magnetic fields to move seawater and brine without mechanical parts. &
The idea is exotic, but it does not directly solve salt separation: it replaces mechanical pumping with electromagnetic pumping. It was rated low feasibility (\(F=1\)) with null structural validity. \\
\addlinespace
Desalination &
\textit{self\_reflection} &
\textbf{Resonant acoustic freeze desalination.} Combines bipolar sub-nanosecond electrical pulses to weaken hydration layers with acoustic ice nucleation via ultrasound. &
Self-reflection produces a much more polished and metrically strong hypothesis (\(F=3\), \(Q=1.000\)), but it retains a significant dependence on active fields and fine acoustic control. It improves the argumentative envelope without fully shifting the mechanism toward an external constraint. \\
\addlinespace
Desalination &
\textit{full\_system} &
\textbf{Cryo-flotation desalination with cyclopentane clathrates.} Uses cyclopentane to form clathrates at \(7\,^\circ\mathrm{C}\), geometrically exclude salt, separate by buoyancy, and couple the cycle to an industrial cryogenic source such as LNG regasification. &
The critique discarded MHD, cavitation, and active fields due to parasitic heating or dielectric breakdown. The result retains the useful core: salt exclusion by clathrates, explicit management of latent heat, and a passive separation mechanism. \\
\addlinespace
Parliamentary deadlock &
\textit{direct\_llm} &
\textbf{Meritocratic delegated voting with zero-knowledge proofs.} Assigns different voting weight to deputies according to a merit score computed by an external oracle and applies a cryptographic penalty to party discipline. &
The proposal introduces a flashy but democratically fragile solution: an unelected oracle alters the voting weight of elected representatives. It was rated low feasibility (\(F=1\)) with null structural validity. \\
\addlinespace
Parliamentary deadlock &
\textit{self\_reflection} &
\textbf{Randomized evidence-prediction tournaments.} Randomly distributes deputies into teams that predict the future validity of legislative evidence and accumulate reputation for accuracy. &
A pragmatic, high-quality output (\(F=3\), \(Q=1.000\)): it could work as an advisory tool in committees. However, it mainly addresses the epistemic quality of deliberation, not the coercive core of party discipline. \\
\addlinespace
Parliamentary deadlock &
\textit{full\_system} &
\textbf{Electoral epistemic deposit with adversarial resolution.} Deputies dissent by betting their future position on electoral lists against exogenous legislative indicators, verified by a multi-party adversarial jury. &
The critical pressure discards secret voting, vote weighting, and unconstitutional mechanisms. The final proposal preserves formal equality of the vote and shifts the incentive to the internal party cost, making epistemological friction more tangible. \\
\bottomrule
\end{tabular}%
}
\caption*{\footnotesize Source: own elaboration based on experimental logs and real JSON outputs. Secondary examples are detailed in Table~\ref{tab:appendix-qualitative-examples-en}. Interpretive reading: \textit{self\_reflection} is especially competitive in conceptual, deliberative, or sociopolitical problems where the main improvement comes from internal coherence and argumentative refinement; \textit{full\_system} seems more appropriate when the hypothesis must survive hard external constraints, such as physical feasibility, thermodynamics, mathematical consistency, legal fit, or empirical contrast.}
\end{table*}

These examples qualify the quantitative reading: \textit{self\_reflection} is not simply a weak version of the full system, but a mode that often improves the coherence and presentation of an initial line of reasoning. The qualitative difference is that the full system introduces additional external friction, through empirical search, adversarial debate, and semantic filtering, which tends to change the surviving mechanism rather than merely refine it. For this reason, the choice of architecture depends on the domain: in philosophical, deliberative, or sociological tasks, self-reflection may suffice to obtain solid proposals; in domains where an attractive idea may fail due to physical, mathematical, legal, or empirical limits, the full system offers a more demanding methodological barrier. Other modes with ablations show different failures and virtues: without search, physically spectacular but poorly grounded solutions reappear, while without lateral lenses more sober but less transdisciplinary institutional proposals emerge. Additional examples are collected in the appendix.

\subsection{Discussion and assessment of the results}
The results show an important tension between structural quality, diversity, and operational cost. All experimental conditions completed the 30 planned runs, so the completion rate does not bias the main comparison. Even so, the full system requires many more inferences than direct prompting: initial configuration, eight rounds of generation, critique, filtering, and final evaluation. Therefore, the comparisons describe the effect of different architectures under the experimental design used, but they do not demonstrate efficiency per token, economic cost, or latency.

The full system achieves greater originality and combinatorial rarity than the direct model and produces more constrained qualitative examples, as shown in Table~\ref{tab:qualitative-examples-en}. However, it does not dominate all the ablations. \textit{self\_reflection} obtains the highest rate of structurally valid, highly novel hypotheses and the lowest collapse rate among evaluable runs, while \textit{no\_search} achieves the highest productivity. This does not invalidate the full architecture, but it does prevent a strong conclusion of general superiority. It is worth noting that, once collapse is restricted to runs with at least two hypotheses, the differences between conditions on this metric are small (between 0.392 and 0.458): much of the apparent contrast with the direct model stems from its low fluency, not from extreme semantic concentration.

The ablations also reveal a clean trade-off in the lateral lenses: removing them improves feasibility in all three blocks (reaching a perfect feasible rate of 1.000), but reduces combinatorial rarity just as systematically (\(\Delta = +0.178\) in favor of the full system across all three blocks). In other words, lateral lenses buy transdisciplinary leaps at a measurable cost in immediate plausibility.

The most prudent interpretation is that each module shifts a different trade-off. External search and adversarial critique tend to increase plausibility pressure; the semantic filter helps convert long critiques into reusable constraints; and lateral lenses favor less frequent combinations. Table~\ref{tab:appendix-per-problem-en} disaggregates the main metrics by problem and supports the domain-dependent reading: in desalination, \textit{self\_reflection} and \textit{no\_lateral\_lenses} reach the highest structural validity rates, while the full system maintains high originality and rarity; in the parliamentary deadlock problem, the full system maximizes originality and combinatorial rarity, but its structural validity drops below the baselines. In physical, technical, or legal problems, that external friction can be valuable; in more deliberative or conceptual problems, a well-designed self-reflection may suffice and be cheaper.

\section{Conclusions}
The results obtained partially support a bounded hypothesis: hallucination in language models does not appear as a useful resource when accepted directly, but when treated as speculative generation subjected to critique, filtering, and evaluation. Therefore, the work does not show that hallucinating is beneficial in itself; it shows that an architecture separating speculation and control can transform some divergent outputs into more structured, debatable, and evaluable hypotheses.

The study shows that direct prompting tends to be one of the weakest conditions across most observed metrics. Its proposals can be striking, but they are often poorly constrained by criteria of feasibility, specificity, or empirical fit. By contrast, the conditions based on reflection or structured architecture significantly modify the type of hypotheses produced. Nonetheless, the results also qualify the initial hypothesis: the full system does not dominate all metrics relative to simple self-reflection. The \textit{self\_reflection} condition is especially competitive, suggesting that in conceptual, deliberative, or sociopolitical problems a well-formulated internal review may be sufficient to produce coherent and useful proposals.

The full architecture seems to provide its greatest value in domains where an attractive idea must survive strong external constraints. In physical, technical, legal, or empirical problems, external search, adversarial critique, and the semantic filter help shift proposals from spectacular but fragile solutions toward more defensible mechanisms. The ablations reinforce this interpretation: removing search favors more diverse or striking but less grounded ideas; removing lateral lenses produces more sober but less transdisciplinary solutions; and removing the semantic filter reduces the system's capacity to convert critique into constructive pressure.

The work has important limitations. Only three problem--seed blocks were used; although there were 180 runs, the paired comparisons remain exploratory because they do not cover 180 independent tasks. Moreover, no normalization was applied for the number of calls, tokens, or cost; part of the improvement may come from a larger inference budget. This constraint must be interpreted in the material context of the study: this is work carried out in the first year of \textit{Bachillerato} (upper secondary education), with limited computational and financial resources, not an experiment run with the infrastructure of an industrial laboratory or a large artificial intelligence company. For this reason, the design prioritized a bounded but traceable matrix of runs on Vertex AI, sufficient to observe internal descriptive trends, but not to reach the statistical or computational scale of organizations with substantially greater research resources. The metrics used evaluate structural validity, originality, feasibility, and non-duplication, but they do not constitute definitive scientific validation. Feasibility depends on an LLM evaluator and novelty is measured with a lexical-semantic proxy, not with expert literature review. The generated hypotheses should be understood as initial candidates for further research, not as proven discoveries.

Overall, these results suggest that useful computational creativity does not require eliminating all hallucinatory behavior, but distinguishing between dangerous and controlled uses of speculation. In informational contexts, hallucination remains a problem that must be mitigated; in research contexts, it can become a source of hypotheses if it is surrounded by epistemological friction, empirical grounding, and explicit evaluation. Hallucination, therefore, is not automatically a resource: it only becomes one when an architecture forces it to transform into provisionally evaluable knowledge.

\section*{Acknowledgments}
I thank the project tutor/coordinator for their guidance during the development of the project and the review of the manuscript. I also thank my father, Fernando Rodríguez Merino, PhD in Physics, for his support and guidance, as well as for his help in funding the experimental runs carried out on Google Cloud Vertex AI.

\section*{Supplementary information}

\subsection*{Additional tables}
\begin{table*}[!tp]
\centering
\scriptsize
\caption{Aggregated results by experimental condition on the main metrics.}
\label{tab:appendix-main-metrics-en}
\resizebox{\textwidth}{!}{%
\begin{tabular}{lcccccc}
\toprule
\textbf{Condition} &
\textbf{Structural validity} &
\textbf{Feasible rate} &
\textbf{Originality} &
\textbf{Comb. rarity} &
\textbf{Diversity} &
\textbf{Collapse} \\
\midrule
\textit{direct\_llm} & 0.411 & 0.678 & 0.389 & 0.491 & 0.776 & 0.417 \\
\textit{self\_reflection} & \textbf{0.656} & 0.950 & 0.622 & 0.830 & 0.790 & \textbf{0.392} \\
\textit{no\_agent4} & 0.500 & 0.967 & 0.607 & 0.757 & 0.733 & 0.458 \\
\textit{no\_search} & 0.411 & 0.936 & 0.651 & 0.745 & \textbf{0.821} & 0.403 \\
\textit{no\_lateral\_lenses} & 0.561 & \textbf{1.000} & 0.647 & 0.670 & 0.818 & 0.432 \\
\textit{full\_system} & 0.444 & 0.944 & \textbf{0.652} & \textbf{0.848} & 0.766 & 0.405 \\
\bottomrule
\end{tabular}%
}
\caption*{\footnotesize Source: own elaboration. Values are normalized between 0 and 1. In all columns higher is better, except for \textit{Collapse}, where lower values indicate less semantic concentration. The feasible rate corresponds to \(FeasibleRate\), not to the ordinal mean of feasibility. Collapse is computed only over runs with at least two hypotheses; for \textit{direct\_llm} only 2 of 30 runs meet this condition, so its value should be read with particular caution.}
\end{table*}

\begin{table*}[!tp]
\centering
\scriptsize
\caption{Computation of the feasible combinatorial creative performance proxy.}
\label{tab:appendix-fccy-en}
\resizebox{0.92\textwidth}{!}{%
\begin{tabular}{lccccc}
\toprule
\textbf{Condition} &
\textbf{Structural validity} &
\textbf{Feasible rate} &
\textbf{Originality} &
\textbf{Comb. rarity} &
\textbf{FCCY index} \\
\midrule
\textit{self\_reflection} & 0.656 & 0.950 & 0.622 & 0.830 & \textbf{0.321} \\
\textit{no\_lateral\_lenses} & 0.561 & 1.000 & 0.647 & 0.670 & 0.243 \\
\textit{full\_system} & 0.444 & 0.944 & 0.652 & 0.848 & 0.232 \\
\textit{no\_agent4} & 0.500 & 0.967 & 0.607 & 0.757 & 0.222 \\
\textit{no\_search} & 0.411 & 0.936 & 0.651 & 0.745 & 0.187 \\
\textit{direct\_llm} & 0.411 & 0.678 & 0.389 & 0.491 & 0.053 \\
\bottomrule
\end{tabular}%
}
\caption*{\footnotesize Source: own elaboration. We define \(FCCY = StructuralValidity \times FeasibleRate \times Originality \times CombinationRarity\). This index does not replace the individual metrics: it only summarizes the production of hypotheses that are simultaneously valid, feasible, original, and combinatorially rare.}
\end{table*}

\begin{table*}[!tp]
\centering
\scriptsize
\caption{Auxiliary metrics of productivity, quality, and composite creativity.}
\label{tab:appendix-auxiliary-metrics-en}
\resizebox{\textwidth}{!}{%
\begin{tabular}{lccccccc}
\toprule
\textbf{Condition} &
\textbf{Fluency} &
\textbf{Quality} &
\textbf{Usefulness} &
\textbf{Creativity} &
\textbf{SVHN rate} &
\textbf{VCN} &
\textbf{Valid rare} \\
\midrule
\textit{direct\_llm} & 1.100 & 0.736 & 0.802 & 0.567 & 0.144 & 0.174 & 0.133 \\
\textit{self\_reflection} & 1.567 & 0.793 & 0.844 & 0.713 & \textbf{0.656} & \textbf{0.534} & \textbf{0.600} \\
\textit{no\_agent4} & 1.500 & 0.778 & 0.833 & 0.696 & 0.467 & 0.343 & 0.367 \\
\textit{no\_search} & \textbf{2.000} & 0.759 & 0.819 & 0.726 & 0.411 & 0.278 & 0.367 \\
\textit{no\_lateral\_lenses} & 1.533 & \textbf{0.793} & \textbf{0.845} & \textbf{0.728} & 0.561 & 0.423 & 0.467 \\
\textit{full\_system} & 1.767 & 0.770 & 0.827 & 0.719 & 0.444 & 0.381 & 0.467 \\
\bottomrule
\end{tabular}%
}
\caption*{\footnotesize Source: own elaboration. Fluency represents the mean number of hypotheses extracted per successful, evaluable run; the SVHN rate summarizes structurally valid, highly novel hypotheses; VCN corresponds to valid combinatorial novelty. The best value in each column is marked in bold.}
\end{table*}

\begin{table*}[!tp]
\centering
\scriptsize
\caption{Completion rate and technical failures by condition.}
\label{tab:appendix-completion-en}
\resizebox{\textwidth}{!}{%
\begin{tabular}{lccccccc}
\toprule
\textbf{Condition} &
\textbf{Expected} &
\textbf{Completed} &
\textbf{Failed} &
\textbf{Compl. rate} &
\textbf{API issues} &
\textbf{API limit} &
\textbf{Other technical} \\
\midrule
\textit{direct\_llm} & 30 & 30 & 0 & \textbf{1.000} & 0.000 & 0.000 & 0.000 \\
\textit{self\_reflection} & 30 & 30 & 0 & \textbf{1.000} & 0.000 & 0.000 & 0.000 \\
\textit{no\_agent4} & 30 & 30 & 0 & \textbf{1.000} & 0.000 & 0.000 & 0.000 \\
\textit{no\_search} & 30 & 30 & 0 & \textbf{1.000} & 0.000 & 0.000 & 0.000 \\
\textit{no\_lateral\_lenses} & 30 & 30 & 0 & \textbf{1.000} & 0.000 & 0.000 & 0.000 \\
\textit{full\_system} & 30 & 30 & 0 & \textbf{1.000} & 0.000 & 0.000 & 0.000 \\
\bottomrule
\end{tabular}%
}
\caption*{\footnotesize Source: own elaboration. The technical \textit{join\_failed} row of the manifest is not part of the experimental design and is excluded from this table.}
\end{table*}

\begin{table*}[!tp]
\centering
\scriptsize
\caption{Paired differences of the full system relative to each baseline.}
\label{tab:appendix-paired-deltas-en}
\resizebox{\textwidth}{!}{%
\begin{tabular}{llccccc}
\toprule
\textbf{Comparator} &
\textbf{Metric} &
\textbf{Blocks} &
\(\boldsymbol{\Delta}\) \textbf{mean} &
\textbf{Win rate} &
\textbf{Cohen \(d_z\)} &
\textbf{95\% CI} \\
\midrule
\textit{direct\_llm} & Feasibility & 3 & 0.139 & 1.000 & 1.981 & [0.075; 0.214] \\
\textit{direct\_llm} & Originality & 3 & 0.263 & 1.000 & 7.896 & [0.235; 0.300] \\
\textit{direct\_llm} & Comb. rarity & 3 & 0.357 & 1.000 & 1.265 & [0.068; 0.631] \\
\textit{direct\_llm} & Diversity & 2 & 0.006 & 0.500 & 0.064 & [-0.056; 0.067] \\
\textit{direct\_llm} & Collapse & 2 & -0.031 & 0.500 & -0.193 & [-0.146; 0.083] \\
\textit{direct\_llm} & Creativity & 3 & 0.153 & 1.000 & 3.774 & [0.108; 0.187] \\
\textit{self\_reflection} & Feasibility & 3 & 0.021 & 0.667 & 0.346 & [-0.044; 0.078] \\
\textit{self\_reflection} & Originality & 3 & 0.031 & 1.000 & 2.033 & [0.013; 0.041] \\
\textit{self\_reflection} & Comb. rarity & 3 & 0.019 & 0.667 & 0.231 & [-0.073; 0.078] \\
\textit{self\_reflection} & Diversity & 3 & -0.005 & 0.667 & -0.113 & [-0.060; 0.025] \\
\textit{self\_reflection} & Collapse & 3 & -0.021 & 0.333 & -0.375 & [-0.071; 0.038] \\
\textit{self\_reflection} & Creativity & 3 & 0.006 & 0.667 & 0.283 & [-0.013; 0.029] \\
\textit{no\_agent4} & Feasibility & 3 & 0.011 & 0.667 & 0.296 & [-0.028; 0.047] \\
\textit{no\_agent4} & Originality & 3 & 0.045 & 1.000 & 0.633 & [0.002; 0.128] \\
\textit{no\_agent4} & Comb. rarity & 3 & 0.091 & 0.667 & 0.605 & [-0.017; 0.264] \\
\textit{no\_agent4} & Diversity & 3 & 0.065 & 1.000 & 0.777 & [0.000; 0.159] \\
\textit{no\_agent4} & Collapse & 3 & -0.046 & 0.000 & -2.639 & [-0.062; -0.028] \\
\textit{no\_agent4} & Creativity & 3 & 0.023 & 1.000 & 0.784 & [0.003; 0.058] \\
\textit{no\_search} & Feasibility & 3 & -0.011 & 0.333 & -0.158 & [-0.086; 0.053] \\
\textit{no\_search} & Originality & 3 & 0.001 & 0.667 & 0.036 & [-0.037; 0.028] \\
\textit{no\_search} & Comb. rarity & 3 & 0.103 & 1.000 & 1.014 & [0.037; 0.220] \\
\textit{no\_search} & Diversity & 3 & -0.031 & 0.333 & -0.291 & [-0.123; 0.087] \\
\textit{no\_search} & Collapse & 3 & -0.002 & 0.667 & -0.042 & [-0.052; 0.026] \\
\textit{no\_search} & Creativity & 3 & -0.007 & 0.333 & -0.189 & [-0.036; 0.033] \\
\textit{no\_lateral\_lenses} & Feasibility & 3 & -0.049 & 0.000 & ---\textsuperscript{*} & [-0.050; -0.047] \\
\textit{no\_lateral\_lenses} & Originality & 3 & 0.006 & 0.333 & 0.090 & [-0.040; 0.079] \\
\textit{no\_lateral\_lenses} & Comb. rarity & 3 & 0.178 & 1.000 & 3.924 & [0.147; 0.230] \\
\textit{no\_lateral\_lenses} & Diversity & 3 & -0.027 & 0.333 & -0.428 & [-0.085; 0.042] \\
\textit{no\_lateral\_lenses} & Collapse & 3 & -0.015 & 0.333 & -0.283 & [-0.062; 0.042] \\
\textit{no\_lateral\_lenses} & Creativity & 3 & -0.009 & 0.333 & -0.223 & [-0.039; 0.034] \\
\bottomrule
\end{tabular}%
}
\caption*{\footnotesize Source: own elaboration. We define \(\Delta = m_{\text{system}} - m_{\text{baseline}}\) within the same problem--seed block. The win rate is always the fraction of blocks with \(\Delta > 0\); for the collapse rate, a low win rate and a negative \(\Delta\) therefore favor the full system, because they indicate less semantic collapse. These values should not be read as strong inferential evidence: with only three effective blocks, and only two for some metrics, \(d_z\), the win rate, and the 95\% CI are exploratory descriptors sensitive to small variations and to near-zero variances. \textsuperscript{*}\(d_z\) is omitted when the between-block variance is nearly zero and the resulting value has no useful interpretation.}
\end{table*}

\begin{table*}[!tp]
\centering
\scriptsize
\caption{Average ranking by condition on selected metrics.}
\label{tab:appendix-average-ranks-en}
\resizebox{\textwidth}{!}{%
\begin{tabular}{lccccccc}
\toprule
\textbf{Condition} &
\textbf{Struct. valid.} &
\textbf{Feasibility} &
\textbf{Originality} &
\textbf{Comb. rarity} &
\textbf{Diversity} &
\textbf{Collapse} &
\textbf{Creativity} \\
\midrule
\textit{direct\_llm} & 4.667 & 6.000 & 6.000 & 5.667 & 4.000 & 3.500 & 6.000 \\
\textit{self\_reflection} & \textbf{1.667} & \textbf{2.333} & 4.000 & 2.000 & \textbf{3.000} & 3.667 & 3.333 \\
\textit{no\_agent4} & 3.000 & 3.333 & 4.000 & 3.000 & 4.333 & 4.667 & 4.333 \\
\textit{no\_search} & 5.000 & 3.667 & 3.000 & 4.000 & \textbf{3.000} & \textbf{2.333} & 2.667 \\
\textit{no\_lateral\_lenses} & 2.667 & \textbf{2.333} & \textbf{2.000} & 4.667 & \textbf{3.000} & 3.333 & \textbf{2.000} \\
\textit{full\_system} & 4.000 & 3.333 & \textbf{2.000} & \textbf{1.667} & \textbf{3.000} & 2.667 & 2.667 \\
\bottomrule
\end{tabular}%
}
\caption*{\footnotesize Source: own elaboration. The ranking is computed within problem--seed blocks; lower values indicate a better relative position. In \textit{Collapse}, a lower ranking is also better because it corresponds to a lower \textit{CollapseRate}; in that metric \textit{direct\_llm} only participates in the 2 blocks where it produced at least two hypotheses. The best average ranking in each metric is marked in bold.}
\end{table*}

\begin{table*}[!tp]
\centering
\scriptsize
\caption{Main metrics disaggregated by experimental problem.}
\label{tab:appendix-per-problem-en}
\resizebox{\textwidth}{!}{%
\begin{tabular}{llcccccc}
\toprule
\textbf{Problem} &
\textbf{Condition} &
\textbf{Structural validity} &
\textbf{Feasible rate} &
\textbf{Originality} &
\textbf{Comb. rarity} &
\textbf{SVHN rate} &
\textbf{Quality} \\
\midrule
Desalination & \textit{direct\_llm} & 0.367 & 0.617 & 0.350 & 0.651 & 0.067 & 0.725 \\
Desalination & \textit{self\_reflection} & \textbf{0.750} & 0.925 & 0.601 & 0.869 & \textbf{0.750} & 0.794 \\
Desalination & \textit{no\_agent4} & 0.483 & 0.975 & 0.623 & 0.866 & 0.433 & 0.780 \\
Desalination & \textit{no\_search} & 0.483 & 0.904 & 0.632 & 0.743 & 0.483 & 0.756 \\
Desalination & \textit{no\_lateral\_lenses} & 0.700 & \textbf{1.000} & \textbf{0.658} & 0.678 & 0.700 & \textbf{0.812} \\
Desalination & \textit{full\_system} & 0.567 & 0.933 & 0.627 & \textbf{0.871} & 0.567 & 0.784 \\
\midrule
Parliament & \textit{direct\_llm} & 0.500 & 0.800 & 0.468 & 0.171 & 0.300 & 0.758 \\
Parliament & \textit{self\_reflection} & 0.467 & \textbf{1.000} & 0.663 & 0.751 & 0.467 & \textbf{0.791} \\
Parliament & \textit{no\_agent4} & \textbf{0.533} & 0.950 & 0.575 & 0.538 & \textbf{0.533} & 0.773 \\
Parliament & \textit{no\_search} & 0.267 & \textbf{1.000} & 0.690 & 0.750 & 0.267 & 0.765 \\
Parliament & \textit{no\_lateral\_lenses} & 0.283 & \textbf{1.000} & 0.625 & 0.656 & 0.283 & 0.756 \\
Parliament & \textit{full\_system} & 0.200 & 0.967 & \textbf{0.704} & \textbf{0.802} & 0.200 & 0.741 \\
\bottomrule
\end{tabular}%
}
\caption*{\footnotesize Source: own elaboration. Per-run means aggregated by problem and condition; desalination combines the seed-42 and seed-43 blocks, parliament corresponds to seed 42. The best value for each metric within each problem is marked in bold. The relatively high structural validity of \textit{direct\_llm} on the parliament problem coexists with the lowest combinatorial rarity of the entire experiment (0.171): it produces conventional proposals that are therefore easy to validate.}
\end{table*}

\subsection*{Examples of system outputs}
Table~\ref{tab:appendix-qualitative-examples-en} extends the traceability of Table~\ref{tab:qualitative-examples-en} and collects additional examples from the ablation conditions. These cases should not be read as an absolute hierarchy, but as evidence of different behavioral modes: some ablations produce very solid solutions, while others clearly reveal the role of empirical grounding or final filtering.

\begin{table*}[!tp]
\centering
\scriptsize
\caption{Provenance and additional qualitative examples by experimental condition.}
\label{tab:appendix-qualitative-examples-en}
\resizebox{\textwidth}{!}{%
\begin{tabular}{p{1.7cm}p{2.2cm}p{3.8cm}p{6.1cm}p{4.0cm}}
\toprule
\textbf{Use} &
\textbf{Condition} &
\textbf{Hypothesis} &
\textbf{Qualitative reading} &
\textbf{Provenance and metrics} \\
\midrule
Main &
\textit{direct\_llm} &
\textbf{Electromagnetic pumping of seawater in the solid state} &
Striking but misdirected idea: it replaces mechanical pumping with MHD without directly solving ionic separation. It serves as an example of attractive speculation with low physical constraint. &
Run 0065, id 3. Desalination, seed 43, repetition 05. \(F=1\), \(Q=0.528\). \\
\addlinespace
Main &
\textit{self\_reflection} &
\textbf{Resonant acoustic freeze desalination} &
Self-reflection greatly improves the final form and reaches a high score, but it retains a strong dependence on sub-nanosecond PEF and precise acoustic nucleation. &
Run 0073, id 2. Desalination, seed 43, repetition 03. \(F=3\), \(Q=1.000\). \\
\addlinespace
Main &
\textit{full\_system} &
\textbf{Cryo-flotation desalination with cyclopentane clathrates} &
External critique shifts the design toward a more passive separation mechanism: cyclopentane clathrates, buoyancy, and explicit management of latent heat. &
Run 0117, id 1. Desalination, seed 43, repetition 07. \(F=3\), \(Q=0.917\). \\
\addlinespace
Main &
\textit{direct\_llm} &
\textbf{Meritocratic delegated voting with zero-knowledge proofs} &
Introduces a merit oracle that alters parliamentary voting weight. The proposal is sophisticated on the surface, but politically fragile and structurally invalid. &
Run 0129, id 1. Parliament, seed 42, repetition 09. \(F=1\), \(Q=0.694\). \\
\addlinespace
Main &
\textit{self\_reflection} &
\textbf{Randomized evidence-prediction tournaments} &
Strong example of self-reflection: it proposes an implementable advisory mechanism to improve evidence evaluation, although it does not directly address party coercion. &
Run 0137, id 3. Parliament, seed 42, repetition 07. \(F=3\), \(Q=1.000\). \\
\addlinespace
Main &
\textit{full\_system} &
\textbf{Electoral epistemic deposit with adversarial resolution} &
The final output avoids weighting votes or using parliamentary secrecy and shifts the incentive to the internal party cost, preserving formal equality of deputies. &
Run 0171, id 1. Parliament, seed 42, repetition 01. \(F=2\), \(Q=0.889\). \\
\midrule
Appendix &
\textit{no\_search} &
\textbf{Supersonic ballistic ion classifier by MHD} &
Without external search, a spectacular MHD solution reappears: fields \(>10\,\mathrm{T}\), oceanic pressure, and Lorentz deflection. It is creative, but very vulnerable to turbulent mixing and physical scale. &
Run 0098, id 1. Desalination, seed 43, repetition 08. \(F=2\), \(Q=0.806\). \\
\addlinespace
Appendix &
\textit{no\_agent4} &
\textbf{Microfluidic polarization of ionic concentration} &
The absence of the last agent does not prevent producing a solid hypothesis: it reduces macroscopic MHD to microfluidic electrokinetics, a physically more defensible route. &
Run 0022, id 3. Desalination, seed 42, repetition 02. \(F=3\), \(Q=0.917\). \\
\addlinespace
Appendix &
\textit{no\_lateral\_lenses} &
\textbf{Electrochemical ion sponge based on Prussian blue analogues} &
Removing lateral lenses yields a less exotic solution, closer to existing electrochemical intercalation technologies. It gains plausibility but loses part of the combinatorial leap. &
Run 0042, id 1. Desalination, seed 42, repetition 02. \(F=3\), \(Q=0.917\). \\
\addlinespace
Appendix &
\textit{no\_search} &
\textbf{Quadratic time market to control the agenda} &
Proposes a procedural economy where the cost of hoarding parliamentary time grows quadratically. It is elegant, although it depends on an institutional formalization not externally contrasted. &
Run 0155, id 1. Parliament, seed 42, repetition 05. \(F=3\), \(Q=0.917\). \\
\addlinespace
Appendix &
\textit{no\_agent4} &
\textbf{Zero-knowledge activation of Article 85.1} &
A particularly valuable example: it uses a local legal anchor, the 70-signature threshold of Article 85.1, and applies cryptography only to protect internal coordination. &
Run 0150, id 1. Parliament, seed 42, repetition 10. \(F=3\), \(Q=0.917\). \\
\addlinespace
Appendix &
\textit{no\_lateral\_lenses} &
\textbf{Consensus-weighted procedural dividend} &
A sober and probably implementable solution: it rewards multi-party consensus with a procedural fast track. It shows that removing laterality can increase institutional clarity. &
Run 0167, id 2. Parliament, seed 42, repetition 07. \(F=3\), \(Q=0.917\). \\
\bottomrule
\end{tabular}%
}
\caption*{\footnotesize Source: own elaboration based on experimental logs and real JSON outputs. \(F\) corresponds to the discrete feasibility scale and \(Q\) to the quality score.}
\end{table*}

\clearpage
\selectlanguage{spanish}
\setcounter{section}{0}
\twocolumn[
\begin{@twocolumnfalse}
\begin{center}
{\LARGE\bfseries Versión en español\par}
\vspace{0.3cm}
{\large\bfseries La alucinación como recurso, no como defecto\par}
\vspace{0.2cm}
{\normalsize Evaluación de una arquitectura multiagente para transformar salidas especulativas de modelos de lenguaje en hipótesis científicas evaluables\par}
\vspace{0.2cm}
{\small Nicolás Rodríguez Álvarez --- IES Parquesol, Valladolid, Castilla y León, España\par}
\end{center}
\vspace{0.5cm}

\section*{Resumen}
Los Grandes Modelos de Lenguaje (LLMs) actuales suelen alinearse para reducir alucinaciones y priorizar la recuperación factual. Aunque esto limita la desinformación, también puede restringir tareas de ideación científica especulativa, donde la creatividad combinatoria depende de asociaciones semánticas no evidentes. En este trabajo se propone una orquestación multiagente implementada en Rust que trata la alucinación como fase controlada de generación de hipótesis, no como resultado final. El sistema introduce un bucle de \textit{fricción epistemológica} entre un agente generador de alta entropía y un agente evaluador apoyado en búsqueda empírica, mediado por un filtro semántico de baja entropía diseñado para reducir ruido y repetición. Los experimentos generaron hipótesis evaluadas por viabilidad en dominios físicos y sociopolíticos. Además, se presenta un estudio exploratorio que compara el sistema completo con una petición directa al modelo, autorreflexión simple y tres ablaciones. Los resultados sitúan la petición directa al modelo entre las condiciones más débiles en la mayoría de métricas observadas, pero no muestran una superioridad general del sistema completo frente a la autorreflexión simple. Sugieren que cada arquitectura modifica de forma distinta el equilibrio entre originalidad, factibilidad, diversidad y anclaje empírico, y que el sistema completo aporta ventajas cuando las hipótesis deben sobrevivir a restricciones físicas, empíricas o institucionales fuertes. En conjunto, el estudio no demuestra que la alucinación sea útil por sí sola; sugiere que la generación especulativa solo gana valor cuando se rodea de restricciones arquitectónicas, anclaje empírico y evaluación explícita.

\smallskip
\noindent\textbf{Palabras clave:} modelos de lenguaje; alucinación; creatividad computacional; sistemas multiagente; generación de hipótesis científicas.
\bigskip
\end{@twocolumnfalse}
]

\section{Introducción}
Como punto de partida conceptual, este trabajo toma una analogía funcional de la literatura sobre inteligencia y cognición: la tensión entre procesos narrativos, asociativos o contemplativos y procesos de control orientados a fines \parencite{nilsson_quest_2009}. La referencia a la Red Neuronal por Defecto (DMN, por sus siglas en inglés) no se usa aquí como hipótesis neurocientífica fuerte, sino como metáfora operativa para separar dos funciones computacionales: exploración semántica amplia y restricción evaluativa.

Con este propósito, proponemos una arquitectura multiagente que traduce esa analogía a un diseño técnico. Un agente generador produce asociaciones de alta entropía; otros módulos introducen crítica empírica, filtrado semántico y evaluación final. El objetivo no es afirmar que los LLMs reproduzcan la cognición humana, sino probar si una separación explícita entre especulación y control mejora la producción de hipótesis estructuradas.

\section{Antecedentes}
Paradójicamente, la trayectoria reciente de la investigación en inteligencia artificial intenta, con todos sus recursos, suprimir una capacidad inherente de los Grandes Modelos de Lenguaje actuales: la propensión a \guillemotleft alucinar\guillemotright \parencite{chen_combating_2024,ji_survey_2023}. Los principales laboratorios perciben este fenómeno fundamentalmente como un vector que agrava la desinformación generada por la IA, al producir respuestas que divergen de la información verídica. Por ello, se han desplegado grandes esfuerzos en materia de alineación, aprendizaje por refuerzo a partir de la retroalimentación humana (RLHF) y un riguroso anclaje empírico para garantizar que los modelos actúen como recuperadores de información fiables y no como motores especulativos \parencite{ji_survey_2023}.

Si bien es innegable que la mitigación de la desinformación es crítica para las aplicaciones dirigidas al público \parencite{chen_combating_2024}, esta optimización implacable de la adherencia factual conlleva un grave coste cognitivo: coarta la capacidad del modelo para la ideación combinatoria divergente y fuera de la distribución. Al tratar todas las alucinaciones de manera uniforme como un \guillemotleft defecto\guillemotright que debe eliminarse, la disciplina neutraliza inadvertidamente el mecanismo mismo que permite a los LLMs realizar asociaciones entre dominios semánticos distantes.

En términos de diseño, puede trazarse una analogía limitada entre generación autorregresiva de alta entropía y pensamiento asociativo: ambas pueden producir conexiones no evidentes, pero también errores, adulación o propuestas físicamente inviables. Por el contrario, una alineación orientada solo a evitar riesgo puede favorecer respuestas más conservadoras. La cuestión central no es si los LLMs poseen un equivalente biológico de la DMN, sino si una arquitectura separada de generación especulativa y control crítico puede aprovechar parte de esa divergencia sin aceptar sus salidas sin verificación.

\section{Hipótesis de trabajo y objetivos de la investigación}

\subsection{Hipótesis de trabajo}
Planteamos una hipótesis arquitectónica, no neurocognitiva: una fase generativa de alta entropía puede aumentar la exploración de hipótesis si se separa explícitamente de una fase posterior de crítica, filtrado y evaluación. La predicción no es que la alucinación sea beneficiosa por sí misma, sino que la especulación controlada puede cambiar el perfil de las hipótesis producidas.

\subsection{Objetivos}
En este artículo, el objetivo principal es presentar un orquestador de I+D multiagente programado en Rust que pone en práctica esta separación funcional. Al forzar la «fricción epistemológica» entre un agente generador especulativo y un agente evaluador empírico basado en búsqueda, buscamos evaluar si una fase de generación divergente, sometida después a crítica y filtrado, produce hipótesis más estructuradas que una petición directa al modelo.

A diferencia de los marcos multiagente estándar orientados a objetivos diseñados para la ejecución de tareas y el uso de herramientas (como ReAct \parencite{yao_react_2023} o AutoGPT \parencite{yang_auto-gpt_2023}), nuestra arquitectura busca optimizarse estrictamente para la fricción teórica especulativa.

\section{Materiales y métodos}

\subsection{Diseño general del sistema}
El sistema desarrollado estudia si la generación especulativa de los modelos de lenguaje puede utilizarse como una fase controlada de ideación científica. El objetivo no es aceptar directamente las salidas no verificadas del modelo, sino tratarlas como hipótesis candidatas que posteriormente deben ser criticadas, filtradas y evaluadas.

La arquitectura se diseñó como un sistema multiagente. En lugar de pedir una única respuesta directa a un LLM, el proceso se divide en funciones diferenciadas: generación de hipótesis, crítica, refinamiento, filtrado semántico y evaluación. Esta decisión se relaciona con trabajos recientes sobre refinamiento iterativo mediante retroalimentación del propio modelo, como \textit{Self-Refine} y \textit{Reflexion}, y con enfoques de debate multiagente orientados a mejorar razonamiento y factualidad \parencite{madaan_self-refine_2023,shinn_reflexion_2023,du_improving_2023}.

La versión completa del sistema incluye varios roles: un módulo que configura el dominio del problema, un agente generador de hipótesis, un agente crítico, un filtro semántico y un evaluador final. La idea central es crear una tensión entre exploración y restricción. A esta tensión la denominamos \textit{fricción epistemológica}: un proceso en el que las ideas especulativas se someten a criterios de viabilidad, especificidad, no duplicación y novedad antes de ser consideradas resultados útiles.

\begin{figure*}[!tp]
    \centering
    \includegraphics[width=0.7\textwidth]{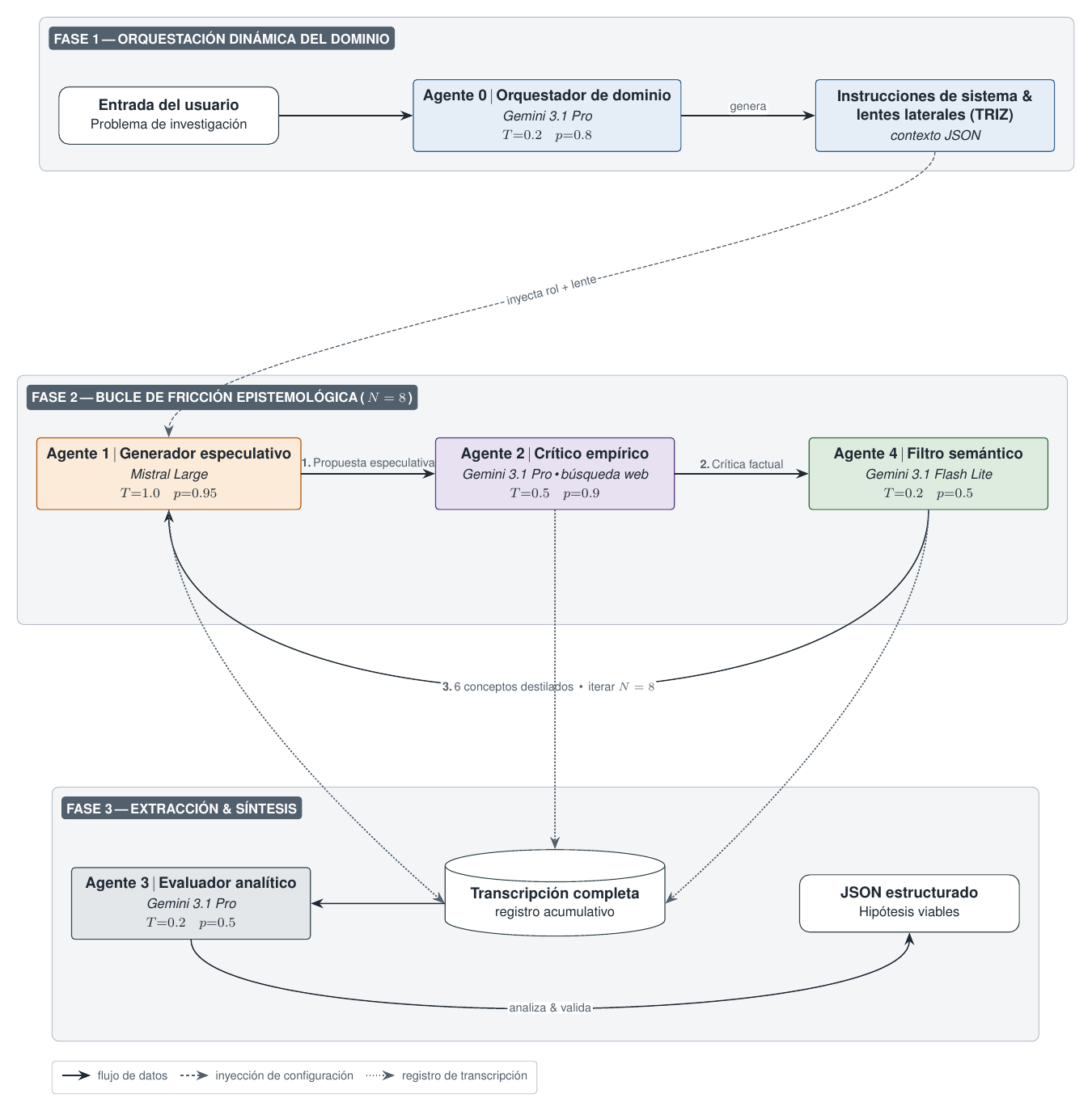}
    \caption{Arquitectura del pipeline multiagente y flujo de datos entre configuración, bucle de fricción epistemológica y evaluación final.}
    \label{fig:pipeline-es}
    \caption*{\footnotesize Fuente/procedencia: elaboración propia a partir del diagrama TikZ y del orquestador implementado para este estudio.}
\end{figure*}

\noindent\textbf{Esquema operativo.} El flujo completo es: (1) entrada del problema y restricciones; (2) configuración automática del dominio, instrucciones del sistema y lentes laterales; (3) generación especulativa por el agente creador; (4) crítica empírica con o sin búsqueda externa, según la condición experimental; (5) filtrado semántico de la crítica y realimentación durante ocho rondas; (6) extracción final de hipótesis en JSON; y (7) cálculo reproducible de métricas sobre los registros experimentales. La Figura~\ref{fig:pipeline-es} resume esta arquitectura.

\subsection{Modelos, herramientas y entorno de ejecución}
El sistema fue implementado principalmente en Rust. La orquestación, las condiciones de ablación, la elección de lentes laterales y la escritura de registros son deterministas para una configuración y semilla dadas; las respuestas concretas de los LLMs siguen siendo estocásticas porque dependen de servicios externos de inferencia. Cada ejecución genera registros estructurados en JSON, transformados después en tablas para su análisis.

Los modelos usados en la ejecución experimental fueron: \path{mistral-large-latest} para el agente generador; \path{gemini-3.1-pro-preview} mediante Vertex AI para configuración de dominio, crítica y evaluación; y \path{gemini-3.1-flash-lite-preview} para el filtro semántico. Los parámetros fueron: generador \(T=1.0\), \(\textit{top-p}=0.95\); crítico \(T=0.5\), \(\textit{top-p}=0.9\); evaluador \(T=0.2\), \(\textit{top-p}=0.5\); arquitecto \(T=0.2\), \(\textit{top-p}=0.8\); extractor/filtro \(T=0.2\), \(\textit{top-p}=0.5\). La búsqueda web de Vertex AI se activó solo en las condiciones con anclaje externo.

El análisis cuantitativo se realizó mediante el paquete Julia \texttt{HallucinationAsAFeatureJulia} \parencite{rodriguez_alvarez_2026_20649714}. A partir de los registros se generaron un manifiesto de ejecuciones, una tabla de hipótesis individuales, métricas agregadas por ejecución, métricas por condición experimental, diferencias pareadas y rankings medios. Esta separación entre ejecución y análisis permite recalcular las métricas sin repetir todos los experimentos generativos.

\subsection{Implementación reproducible de las métricas}
La factibilidad no fue anotada manualmente. La asignó el evaluador final (Agente 3) mediante una salida JSON obligatoria con el campo \texttt{feasibility\_rank}. La conversión fue determinista: \textit{low}=1, \textit{medium}=2 y \textit{high}=3; valores numéricos, cuando aparecían, se acotaron al intervalo 0--3. En términos operativos, una puntuación 1 indica una idea con barreras técnicas o institucionales dominantes; 2, una idea plausible pero con riesgos importantes de implementación o validación; y 3, una idea con mecanismo claro y ruta de prueba razonable. Esta decisión introduce circularidad parcial porque un LLM evalúa salidas de LLMs; por ello la factibilidad se interpreta como puntuación de rúbrica y no como validación experta independiente. La validación científica estricta requeriría revisión por especialistas, experimentos o contraste con conjuntos de datos externos.

Las variables de especificidad se calcularon de forma determinista. El dominio se hereda del problema experimental; la presencia de mecanismo, variable medible y testabilidad se detecta primero mediante campos explícitos si existen y, en su defecto, mediante patrones léxicos fijos. Por ejemplo, el mecanismo se infiere a partir de familias léxicas asociadas a vía causal, mecanismo, efecto, uso o gradiente; la variable medible a partir de magnitudes, unidades, rendimiento, temperatura, presión, energía, precisión o umbrales; y la testabilidad a partir de términos relacionados con prueba, medida, experimento, simulación, comparación, observabilidad, conjunto de datos o modelo.

La distancia entre hipótesis no se calculó con un modelo externo de embeddings, sino con vectores dispersos de conteo de tokens y similitud coseno. Los tokens se normalizaron a minúsculas y se conservaron palabras alfanuméricas de tres o más caracteres. Por tanto, la métrica debe leerse como un proxy léxico-semántico reproducible, no como una medida profunda de novedad científica.

\subsection{Diseño experimental}
El estudio se planteó como una comparación entre el sistema completo, varias referencias y distintas ablaciones. Una ablación consiste en eliminar o modificar un componente del sistema para observar cómo cambia el comportamiento global. Esto permite estimar qué aporta cada módulo a la generación de hipótesis.

Se compararon seis condiciones experimentales:

\begin{itemize}[leftmargin=*]
    \item \textit{direct\_llm}: respuesta directa de un modelo de lenguaje sin arquitectura multiagente.
    \item \textit{self\_reflection}: generación y revisión simple por el propio modelo, inspirada en métodos de refinamiento iterativo como \textit{Self-Refine} y \textit{Reflexion} \parencite{madaan_self-refine_2023,shinn_reflexion_2023}.
    \item \textit{no\_agent4}: sistema sin filtro semántico.
    \item \textit{no\_search}: sistema sin fase de búsqueda o crítica empírica externa.
    \item \textit{no\_lateral\_lenses}: sistema sin perspectivas laterales alternativas.
    \item \textit{full\_system}: arquitectura multiagente completa.
\end{itemize}

La matriz experimental prevista fue:

\[
\begin{aligned}
& 3 \text{ bloques problema--semilla} \times 6 \text{ condiciones} \\
&\quad \times 10 \text{ repeticiones} = 180 \text{ ejecuciones esperadas}.
\end{aligned}
\]

Los bloques fueron: desalinización con semilla 42, desalinización con semilla 43 y bloqueo parlamentario con semilla 42. Las diez repeticiones por condición reducen la dependencia de una única salida generativa, pero no convierten el estudio en 180 problemas independientes. Además, al evaluar distintas condiciones sobre los mismos bloques problema--semilla, se pueden realizar comparaciones pareadas. La unidad primaria de análisis descriptivo es la ejecución completa; las comparaciones pareadas, al agregarse por bloque problema--semilla, siguen siendo exploratorias porque solo hay tres bloques efectivos.

\subsection{Métricas de evaluación}
La evaluación se diseñó como un marco multidimensional. No se emplea una única puntuación de creatividad, porque una hipótesis puede ser novedosa pero inviable, o factible pero trivial. Esta decisión se apoya en benchmarks recientes como \textit{LiveIdeaBench}, que evalúa generación de ideas científicas mediante dimensiones como originalidad, factibilidad, fluidez, flexibilidad y claridad, y \textit{CreativityPrism}, que separa la creatividad de los LLMs en calidad, novedad y diversidad \parencite{ruan_evaluating_2026,hou_creativityprism_2026}.

Sea \(H_r\) el conjunto de hipótesis generadas en una ejecución \(r\), y sea \(i\) una hipótesis individual. La productividad o fluidez se define como:

\[
Fluidez_r = |H_r|.
\]

La factibilidad se codificó como una puntuación ordinal normalizada:

\[
FeasibilityNorm_i = \frac{Feasibility_i}{3}.
\]

La puntuación \(Feasibility_i\) procede del campo \texttt{feasibility\_rank} asignado por el Agente 3 y transformado de forma fija: \(\textit{low}=1\), \(\textit{medium}=2\), \(\textit{high}=3\).

Se considera factible una hipótesis con puntuación igual o superior a 2:

\[
Feasible_i = \mathds{1}[Feasibility_i \geq 2].
\]

La especificidad mide si la hipótesis contiene cuatro elementos mínimos: dominio, mecanismo, variable medible y posibilidad de prueba. Se calcula como:

\[
\begin{aligned}
Specificity_i = \tfrac{1}{4}\big(
& Domain_i + Mechanism_i \\
&+ MeasurableVariable_i + Testability_i \big).
\end{aligned}
\]

Una hipótesis se considera específica cuando:

\[
Specific_i = \mathds{1}[Specificity_i \geq 0.75].
\]

También se evalúa si la hipótesis es lógicamente posible y si no es duplicada. La no duplicación se estima mediante similitud semántica entre hipótesis de la misma ejecución:

\[
NotDuplicate_i = \mathds{1}[MaxSimilarity_i < 0.90].
\]

En esta implementación, \(MaxSimilarity_i\) es la máxima similitud coseno entre el vector de conteos de tokens de la hipótesis \(i\) y el de cualquier otra hipótesis del mismo run.

Con estos criterios se define la validez estructural:

\[
\begin{aligned}
StructuralValid_i = {}
& Specific_i \land Feasible_i \\
&\land LogicalPossible_i \land NotDuplicate_i.
\end{aligned}
\]

La originalidad se estimó mediante distancia léxico-semántica respecto a las referencias. Sea \(v_i\) el vector de conteos de tokens de la hipótesis \(i\). La distancia usada en todas las métricas es:

\[
d(i,j)=1-\cos(v_i,v_j).
\]

Para cada hipótesis se calcula su distancia al vecino más cercano generado por las condiciones de referencia:

\[
NBN_i = \min_{j \in B_p} d(i,j),
\]

donde \(B_p\) es el conjunto de hipótesis de \textit{direct\_llm} y \textit{self\_reflection} asociadas al mismo problema \(p\), excluyendo el propio archivo cuando procede.

Para evitar un umbral completamente arbitrario, la alta novedad se definió en relación con el cuartil superior de la distribución de novedad del modelo directo:

\[
HighNovel_i = \mathds{1}\!\left[ NBN_i \geq Q_{0.75}(NBN_{direct\_llm}) \right].
\]

La métrica principal del análisis fue el número de hipótesis estructuralmente válidas y de alta novedad por ejecución:

\[
\begin{aligned}
SVHN_r = \sum_{i \in H_r} \mathds{1}\big[\,
& StructuralValid_i \\
&\land HighNovel_i \,\big].
\end{aligned}
\]

Esta métrica mide cuántas hipótesis de un run son simultáneamente específicas, factibles, lógicamente posibles, no redundantes y alejadas de las condiciones de referencia según el proxy léxico-semántico usado.

También se calculó la rareza combinatoria:

\[
CombinationRarity_i = -\log(p(t_i)),
\]

donde \(t_i\) representa la combinación de problema y disciplinas asociadas a la hipótesis. En el análisis se aplica suavizado aditivo, \(p(t_i)=(count(t_i)+1)/(N+K)\), y después se normaliza el resultado al intervalo \([0,1]\) según el mínimo y máximo observados. Las combinaciones menos frecuentes dentro del conjunto experimental reciben valores mayores.

La diversidad semántica se calculó como la distancia media entre pares de hipótesis dentro de una ejecución:

\[
SemanticDiversity_r =
\frac{2}{n_r(n_r-1)}
\sum_{i<j} d(i,j).
\]

Esta métrica solo se calcula cuando una ejecución contiene al menos dos hipótesis. Finalmente, el colapso semántico se estimó como la proporción ocupada por el mayor cluster de hipótesis. Los clusters se construyeron como componentes conexas: dos hipótesis se unen si su similitud coseno es igual o superior a 0,90.

\[
CollapseRate_r =
\frac{\text{tamaño del mayor cluster}}{n_r}.
\]

Valores altos indican que muchas hipótesis se concentran en una misma idea; valores bajos indican mayor dispersión conceptual. Al igual que la diversidad, el colapso solo se calcula para ejecuciones con al menos dos hipótesis: con una única hipótesis el mayor cluster coincide trivialmente con toda la ejecución y la métrica confundiría baja productividad con concentración semántica.

\subsection{Procedimiento de análisis}
El análisis se realizó en cuatro fases. Primero, se creó un manifiesto de ejecuciones con el problema, semilla, condición, archivo de salida y estado de cada run. Esto permitió calcular tasas de finalización y separar fallos técnicos de resultados semánticos.

Segundo, se extrajeron las hipótesis individuales de los registros experimentales y se construyó una tabla en la que cada fila corresponde a una hipótesis. Sobre esta tabla se calcularon las métricas de factibilidad, especificidad, novedad, rareza combinatoria, diversidad y no duplicación.

Tercero, las métricas individuales se agregaron por ejecución y después por condición experimental. Así se obtuvieron valores medios por condición para comparar el sistema completo con las referencias y las ablaciones.

Cuarto, se calcularon comparaciones pareadas. Para una métrica \(m\), la diferencia entre el sistema completo y una referencia dentro del mismo bloque problema--semilla se define como:

\[
\Delta_{b,m} =
m_{\text{sistema},b} - m_{\text{referencia},b}.
\]

Este análisis es más informativo que comparar solo medias globales, porque controla parcialmente la dificultad del problema y la variabilidad introducida por la semilla. Sin embargo, debe interpretarse de forma estrictamente descriptiva: aunque hubo 180 ejecuciones, las diferencias pareadas se agregan por bloque problema--semilla y solo existen tres bloques efectivos. Por tanto, los tamaños de efecto, tasas de victoria e intervalos de confianza del apéndice se incluyen como indicadores exploratorios de dirección y estabilidad interna, no como inferencia estadística robusta ni como prueba de significación poblacional.

En esta versión no se realizó una verificación completa de anclaje basada en afirmaciones para todas las hipótesis. Por ello, la métrica principal se denomina validez estructural, no validación científica definitiva.

\section{Resultados}

\subsection{Resultados cuantitativos}
Los resultados cuantitativos se resumen mediante dos visualizaciones principales. La primera muestra un índice compuesto de rendimiento creativo combinatorio factible, utilizado como proxy para comparar la capacidad de cada condición de producir hipótesis novedosas, factibles y combinatoriamente ricas. La segunda compara las condiciones en varias dimensiones normalizadas: validez estructural, factibilidad, originalidad, rareza combinatoria, diversidad y colapso semántico.

\begin{figure}[!htbp]
    \centering
    \includegraphics[width=\linewidth]{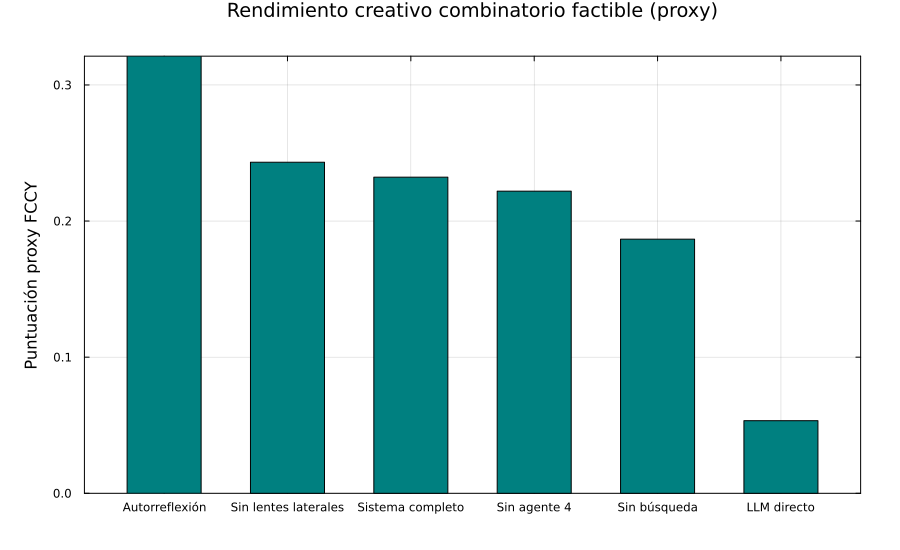}
    \caption{Rendimiento creativo combinatorio factible por condición experimental. El índice FCCY se interpreta como una medida proxy de la capacidad de cada arquitectura para producir hipótesis que combinan factibilidad, originalidad y rareza combinatoria.}
    \label{fig:fccy-es}
    \caption*{\footnotesize Fuente/procedencia: elaboración propia.}
\end{figure}

Como se observa en la Figura~\ref{fig:fccy-es}, las condiciones basadas en reflexión o arquitectura estructurada superan claramente al modelo directo. Este resultado sugiere que introducir procesos de revisión, filtrado o generación estructurada puede aumentar el rendimiento creativo medido por el índice compuesto. No obstante, este índice debe interpretarse como una medida auxiliar, no como una validación científica definitiva de las hipótesis generadas.

\begin{figure*}[!tp]
    \centering
    \includegraphics[width=0.95\textwidth]{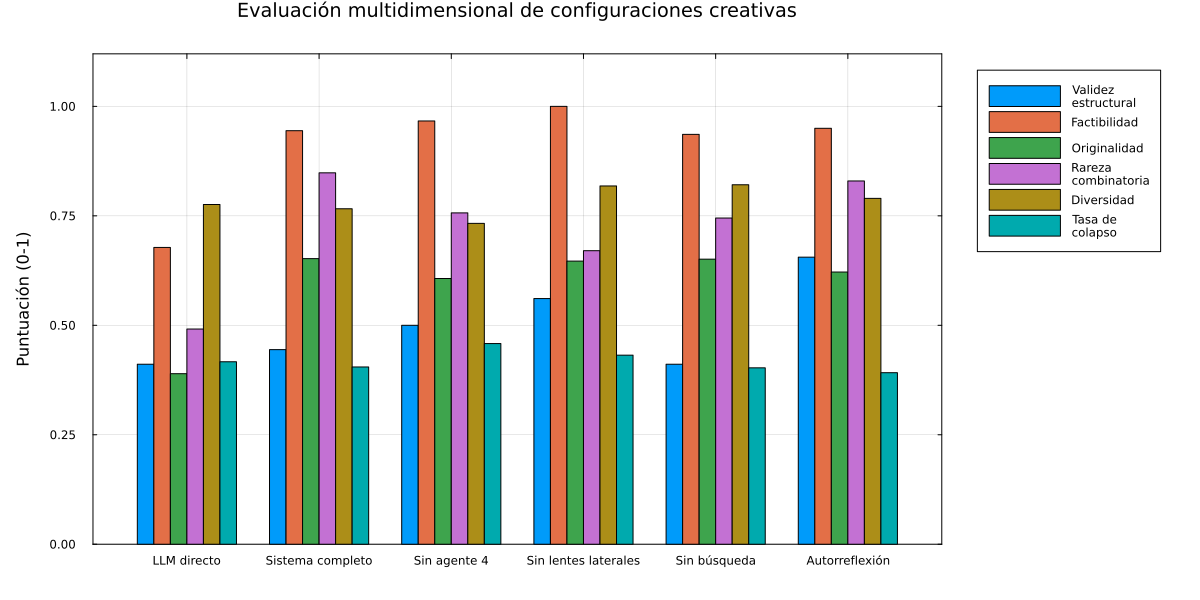}
    \caption{Evaluación multidimensional de las condiciones experimentales. Se comparan seis dimensiones: validez estructural, factibilidad, originalidad, rareza combinatoria, diversidad y tasa de colapso semántico (esta última calculada solo sobre ejecuciones con al menos dos hipótesis).}
    \label{fig:multidimensional-results-es}
    \caption*{\footnotesize Fuente/procedencia: elaboración propia.}
\end{figure*}

La Figura~\ref{fig:multidimensional-results-es} muestra que ninguna condición domina de forma absoluta todas las métricas. El modelo directo obtiene resultados inferiores en la mayoría de dimensiones, mientras que las arquitecturas con reflexión o ablaciones estructuradas muestran mejoras en originalidad, rareza combinatoria y diversidad. Este patrón apoya parcialmente la hipótesis inicial: la generación especulativa puede ser útil como fase de exploración, pero su valor depende de filtros posteriores de factibilidad, especificidad y no redundancia.

Los valores numéricos completos correspondientes a estas figuras se recogen en las tablas adicionales del apéndice.

\subsection{Resultados cualitativos}
Además de las métricas agregadas, se revisaron manualmente salidas reales procedentes de los mismos experimentos. La Tabla~\ref{tab:qualitative-examples-es} muestra contrastes representativos entre el modelo directo, la autorreflexión simple y el sistema completo. En ambos problemas, el modelo directo tiende a producir una idea llamativa pero débilmente constreñida; la autorreflexión mejora la forma y la plausibilidad local, mientras que el sistema completo introduce una presión crítica más externa que desplaza la propuesta hacia mecanismos más específicos o institucionalmente defendibles.

\begin{table*}[!tp]
\centering
\scriptsize
\caption{Ejemplos cualitativos reales de fricción epistemológica entre \textit{direct\_llm}, \textit{self\_reflection} y \textit{full\_system}.}
\label{tab:qualitative-examples-es}
\resizebox{\textwidth}{!}{%
\begin{tabular}{p{2.4cm}p{2.5cm}p{6.4cm}p{5.8cm}}
\toprule
\textbf{Problema} &
\textbf{Condición} &
\textbf{Hipótesis generada} &
\textbf{Lectura cualitativa} \\
\midrule
Desalinización &
\textit{direct\_llm} &
\textbf{Bombeo electromagnético de agua de mar en estado sólido.} Propone usar campos eléctricos y magnéticos cruzados (MHD) o campos magnéticos viajeros para mover agua de mar y salmuera sin piezas mecánicas. &
La idea es exótica, pero no resuelve directamente la separación de sal: sustituye el bombeo mecánico por bombeo electromagnético. Fue evaluada con baja factibilidad (\(F=1\)) y validez estructural nula. \\
\addlinespace
Desalinización &
\textit{self\_reflection} &
\textbf{Desalinización por congelación acústica resonante.} Combina pulsos eléctricos bipolares subnanosegundo para debilitar capas de hidratación con nucleación acústica de hielo mediante ultrasonidos. &
La autorreflexión produce una hipótesis mucho más pulida y métricamente fuerte (\(F=3\), \(Q=1{,}000\)), pero conserva una dependencia importante de campos activos y control acústico fino. Mejora el envoltorio argumental sin desplazar completamente el mecanismo hacia una restricción externa. \\
\addlinespace
Desalinización &
\textit{full\_system} &
\textbf{Desalinización crioflotante con clatratos de ciclopentano.} Usa ciclopentano para formar clatratos a \(7\,^\circ\mathrm{C}\), excluir geométricamente la sal, separar por flotabilidad y acoplar el ciclo a una fuente criogénica industrial como regasificación de GNL. &
La crítica descartó MHD, cavitación y campos activos por calentamiento parásito o ruptura dieléctrica. El resultado conserva el núcleo útil: exclusión salina por clatratos, gestión explícita del calor latente y mecanismo pasivo de separación. \\
\addlinespace
Bloqueo parlamentario &
\textit{direct\_llm} &
\textbf{Voto delegado meritocrático con pruebas de conocimiento cero.} Asigna distinto peso de voto a los diputados según una puntuación de mérito calculada por un oráculo externo y aplica una penalización criptográfica a la disciplina de partido. &
La propuesta introduce una solución vistosa, pero democráticamente frágil: un oráculo no electo altera el peso del voto de representantes electos. Fue evaluada con baja factibilidad (\(F=1\)) y validez estructural nula. \\
\addlinespace
Bloqueo parlamentario &
\textit{self\_reflection} &
\textbf{Torneos aleatorizados de predicción de evidencia.} Distribuye aleatoriamente a diputados en equipos que predicen la validez futura de la evidencia legislativa y acumulan reputación por acierto. &
Es una salida pragmática y de alta calidad (\(F=3\), \(Q=1{,}000\)): podría funcionar como herramienta asesora en comités. Sin embargo, ataca sobre todo la calidad epistémica de la deliberación, no el núcleo coercitivo de la disciplina de partido. \\
\addlinespace
Bloqueo parlamentario &
\textit{full\_system} &
\textbf{Depósito epistémico electoral con resolución adversarial.} Los diputados disienten apostando su futura posición en listas electorales contra indicadores legislativos exógenos, verificados por un jurado adversarial multipartidista. &
La presión crítica descarta voto secreto, ponderación de votos y mecanismos inconstitucionales. La propuesta final mantiene igualdad formal del voto y desplaza el incentivo al coste interno de partido, haciendo más tangible la fricción epistemológica. \\
\bottomrule
\end{tabular}%
}
\caption*{\footnotesize Fuente/procedencia: elaboración propia a partir de registros experimentales y salidas JSON reales. Los ejemplos secundarios se detallan en la Tabla~\ref{tab:appendix-qualitative-examples-es}. Lectura interpretativa: \textit{self\_reflection} resulta especialmente competitivo en problemas conceptuales, deliberativos o sociopolíticos donde la mejora principal procede de coherencia interna y refinamiento argumental; \textit{full\_system} parece más apropiado cuando la hipótesis debe sobrevivir a restricciones externas duras, como factibilidad física, termodinámica, consistencia matemática, encaje jurídico o contraste empírico.}
\end{table*}

Estos ejemplos matizan la lectura cuantitativa: \textit{self\_reflection} no es simplemente una versión débil del sistema completo, sino un modo que a menudo mejora la coherencia y la presentación de una línea de razonamiento inicial. La diferencia cualitativa es que el sistema completo introduce fricción externa adicional, por búsqueda empírica, debate adversarial y filtrado semántico, lo que tiende a cambiar el mecanismo superviviente en lugar de solo refinarlo. Por ello, la elección de arquitectura depende del dominio: en tareas filosóficas, deliberativas o sociológicas, la autorreflexión puede bastar para obtener propuestas sólidas; en dominios donde una idea atractiva puede fallar por límites físicos, matemáticos, jurídicos o empíricos, el sistema completo ofrece una barrera metodológica más exigente. Otros modos con ablaciones muestran fallos y virtudes diferentes: sin búsqueda reaparecen soluciones físicamente espectaculares pero poco ancladas, mientras que sin lentes laterales emergen propuestas institucionales más sobrias aunque menos transdisciplinares. Se recogen ejemplos adicionales en el apéndice.

\subsection{Discusión y valoración de los resultados}
Los resultados muestran una tensión importante entre calidad estructural, diversidad y coste operacional. Todas las condiciones experimentales completaron las 30 ejecuciones previstas, por lo que la tasa de finalización no sesga la comparación principal. Aun así, el sistema completo requiere muchas más inferencias que la petición directa al modelo: configuración inicial, ocho rondas de generación, crítica, filtrado y evaluación final. Por tanto, las comparaciones describen el efecto de distintas arquitecturas bajo el diseño experimental usado, pero no demuestran eficiencia por token, coste económico o latencia.

El sistema completo alcanza mayor originalidad y rareza combinatoria que el modelo directo y produce ejemplos cualitativos más constreñidos, como se observa en la Tabla~\ref{tab:qualitative-examples-es}. Sin embargo, no domina a todas las ablaciones. \textit{self\_reflection} obtiene la mayor tasa de hipótesis estructuralmente válidas de alta novedad y la menor tasa de colapso entre las ejecuciones evaluables, mientras que \textit{no\_search} alcanza la mayor productividad. Esto no invalida la arquitectura completa, pero sí impide una conclusión fuerte de superioridad general. Conviene señalar que, al restringir el colapso a ejecuciones con al menos dos hipótesis, las diferencias entre condiciones en esta métrica son pequeñas (entre 0,392 y 0,458): gran parte del contraste aparente con el modelo directo procede de su baja fluidez, no de una concentración semántica extrema.

Las ablaciones revelan además un compromiso nítido en las lentes laterales: eliminarlas mejora la factibilidad en los tres bloques (alcanzando una tasa factible perfecta de 1,000), pero reduce la rareza combinatoria de forma igualmente sistemática (\(\Delta = +0{,}178\) a favor del sistema completo en los tres bloques). Es decir, las lentes laterales compran salto transdisciplinar pagando un coste medible en plausibilidad inmediata.

La interpretación más prudente es que cada módulo desplaza un compromiso distinto. La búsqueda externa y la crítica adversarial tienden a aumentar la presión de plausibilidad; el filtro semántico ayuda a convertir crítica larga en restricciones reutilizables; y las lentes laterales favorecen combinaciones menos frecuentes. La Tabla~\ref{tab:appendix-per-problem-es} desagrega las métricas principales por problema y apoya la lectura dependiente del dominio: en desalinización, \textit{self\_reflection} y \textit{no\_lateral\_lenses} alcanzan las mayores tasas de validez estructural, mientras que el sistema completo mantiene originalidad y rareza altas; en el bloqueo parlamentario, el sistema completo maximiza originalidad y rareza combinatoria, pero su validez estructural cae por debajo de las referencias. En problemas físicos, técnicos o jurídicos, esa fricción externa puede ser valiosa; en problemas más deliberativos o conceptuales, una autorreflexión bien diseñada puede bastar y resultar más barata.

\section{Conclusiones}
Los resultados obtenidos apoyan parcialmente una hipótesis acotada: la alucinación de los modelos de lenguaje no aparece como recurso útil cuando se acepta directamente, sino cuando se trata como generación especulativa sometida a crítica, filtrado y evaluación. Por tanto, el trabajo no demuestra que alucinar sea beneficioso en sí mismo; muestra que una arquitectura que separa especulación y control puede transformar algunas salidas divergentes en hipótesis más estructuradas, discutibles y evaluables.

El estudio muestra que la petición directa al modelo tiende a ser una de las condiciones más débiles en la mayoría de métricas observadas. Sus propuestas pueden ser llamativas, pero con frecuencia quedan poco constreñidas por criterios de factibilidad, especificidad o encaje empírico. Frente a ello, las condiciones basadas en reflexión o arquitectura estructurada modifican de forma significativa el tipo de hipótesis producidas. No obstante, los resultados también matizan la hipótesis inicial: el sistema completo no domina todas las métricas frente a la autorreflexión simple. La condición \textit{self\_reflection} resulta especialmente competitiva, lo que sugiere que en problemas conceptuales, deliberativos o sociopolíticos una revisión interna bien formulada puede ser suficiente para producir propuestas coherentes y útiles.

La arquitectura completa parece aportar su mayor valor en dominios donde una idea atractiva debe sobrevivir a restricciones externas fuertes. En problemas físicos, técnicos, jurídicos o empíricos, la búsqueda externa, la crítica adversarial y el filtro semántico ayudan a desplazar las propuestas desde soluciones espectaculares pero frágiles hacia mecanismos más defendibles. Las ablaciones refuerzan esta interpretación: eliminar la búsqueda favorece ideas más diversas o llamativas, pero menos ancladas; eliminar las lentes laterales produce soluciones más sobrias, aunque menos transdisciplinares; y eliminar el filtro semántico reduce la capacidad del sistema para convertir la crítica en presión constructiva.

El trabajo presenta limitaciones importantes. Solo se usaron tres bloques problema--semilla; aunque hubo 180 ejecuciones, las comparaciones pareadas siguen siendo exploratorias porque no cubren 180 tareas independientes. Además, no se normalizó por número de llamadas, tokens o coste; parte de la mejora puede proceder de mayor presupuesto inferencial. Esta restricción debe interpretarse en el contexto material del estudio: se trata de un trabajo realizado en 1.º de Bachillerato, con recursos computacionales y económicos limitados, no de un experimento ejecutado con la infraestructura de un laboratorio industrial o de una gran empresa de inteligencia artificial. Por ello, el diseño priorizó una matriz acotada pero trazable de ejecuciones en Vertex AI, suficiente para observar tendencias descriptivas internas, pero no para alcanzar la escala estadística o computacional de organizaciones con recursos de investigación sustancialmente superiores. Las métricas utilizadas evalúan validez estructural, originalidad, factibilidad y no duplicación, pero no constituyen una validación científica definitiva. La factibilidad depende de un evaluador LLM y la novedad se mide con un proxy léxico-semántico, no con revisión experta de literatura. Las hipótesis generadas deben entenderse como candidatos iniciales para investigación posterior, no como descubrimientos comprobados.

En conjunto, estos resultados sugieren que la creatividad computacional útil no requiere eliminar toda conducta alucinatoria, sino distinguir entre usos peligrosos y usos controlados de la especulación. En contextos informativos, la alucinación sigue siendo un problema que debe mitigarse; en contextos de investigación, puede convertirse en una fuente de hipótesis si se rodea de fricción epistemológica, anclaje empírico y evaluación explícita. La alucinación, por tanto, no es automáticamente un recurso: solo llega a serlo cuando una arquitectura la obliga a transformarse en conocimiento provisionalmente evaluable.

\section*{Agradecimientos}
Agradezco al tutor/coordinador del trabajo su orientación durante el desarrollo del proyecto y la revisión del manuscrito. Agradezco también a mi padre, Fernando Rodríguez Merino, doctor en Física, su apoyo y orientación, así como su ayuda para financiar las ejecuciones experimentales realizadas en Google Cloud Vertex AI.

\section*{Información complementaria}

\subsection*{Tablas adicionales}
\begin{table*}[!tp]
\centering
\scriptsize
\caption{Resultados agregados por condición experimental en las métricas principales.}
\label{tab:appendix-main-metrics-es}
\resizebox{\textwidth}{!}{%
\begin{tabular}{lcccccc}
\toprule
\textbf{Condición} &
\textbf{Validez estructural} &
\textbf{Tasa factible} &
\textbf{Originalidad} &
\textbf{Rareza comb.} &
\textbf{Diversidad} &
\textbf{Colapso} \\
\midrule
\textit{direct\_llm} & 0,411 & 0,678 & 0,389 & 0,491 & 0,776 & 0,417 \\
\textit{self\_reflection} & \textbf{0,656} & 0,950 & 0,622 & 0,830 & 0,790 & \textbf{0,392} \\
\textit{no\_agent4} & 0,500 & 0,967 & 0,607 & 0,757 & 0,733 & 0,458 \\
\textit{no\_search} & 0,411 & 0,936 & 0,651 & 0,745 & \textbf{0,821} & 0,403 \\
\textit{no\_lateral\_lenses} & 0,561 & \textbf{1,000} & 0,647 & 0,670 & 0,818 & 0,432 \\
\textit{full\_system} & 0,444 & 0,944 & \textbf{0,652} & \textbf{0,848} & 0,766 & 0,405 \\
\bottomrule
\end{tabular}%
}
\caption*{\footnotesize Fuente/procedencia: elaboración propia. Los valores están normalizados entre 0 y 1. En todas las columnas valores mayores son mejores, excepto en \textit{Colapso}, donde valores menores indican menor concentración semántica. La tasa factible corresponde a \(FeasibleRate\), no a la media ordinal de factibilidad. El colapso se calcula únicamente sobre ejecuciones con al menos dos hipótesis; en \textit{direct\_llm} solo 2 de 30 ejecuciones cumplen esa condición, por lo que su valor debe leerse con especial cautela.}
\end{table*}

\begin{table*}[!tp]
\centering
\scriptsize
\caption{Cálculo del proxy de rendimiento creativo combinatorio factible.}
\label{tab:appendix-fccy-es}
\resizebox{0.92\textwidth}{!}{%
\begin{tabular}{lccccc}
\toprule
\textbf{Condición} &
\textbf{Validez estructural} &
\textbf{Tasa factible} &
\textbf{Originalidad} &
\textbf{Rareza comb.} &
\textbf{Índice FCCY} \\
\midrule
\textit{self\_reflection} & 0,656 & 0,950 & 0,622 & 0,830 & \textbf{0,321} \\
\textit{no\_lateral\_lenses} & 0,561 & 1,000 & 0,647 & 0,670 & 0,243 \\
\textit{full\_system} & 0,444 & 0,944 & 0,652 & 0,848 & 0,232 \\
\textit{no\_agent4} & 0,500 & 0,967 & 0,607 & 0,757 & 0,222 \\
\textit{no\_search} & 0,411 & 0,936 & 0,651 & 0,745 & 0,187 \\
\textit{direct\_llm} & 0,411 & 0,678 & 0,389 & 0,491 & 0,053 \\
\bottomrule
\end{tabular}%
}
\caption*{\footnotesize Fuente/procedencia: elaboración propia. Se define \(FCCY = StructuralValidity \times FeasibleRate \times Originality \times CombinationRarity\). Este índice no sustituye a las métricas individuales: solo resume la producción de hipótesis simultáneamente válidas, factibles, originales y combinatoriamente raras.}
\end{table*}

\begin{table*}[!tp]
\centering
\scriptsize
\caption{Métricas auxiliares de productividad, calidad y creatividad compuesta.}
\label{tab:appendix-auxiliary-metrics-es}
\resizebox{\textwidth}{!}{%
\begin{tabular}{lccccccc}
\toprule
\textbf{Condición} &
\textbf{Fluidez} &
\textbf{Calidad} &
\textbf{Utilidad} &
\textbf{Creatividad} &
\textbf{Tasa SVHN} &
\textbf{VCN} &
\textbf{Válidas raras} \\
\midrule
\textit{direct\_llm} & 1,100 & 0,736 & 0,802 & 0,567 & 0,144 & 0,174 & 0,133 \\
\textit{self\_reflection} & 1,567 & 0,793 & 0,844 & 0,713 & \textbf{0,656} & \textbf{0,534} & \textbf{0,600} \\
\textit{no\_agent4} & 1,500 & 0,778 & 0,833 & 0,696 & 0,467 & 0,343 & 0,367 \\
\textit{no\_search} & \textbf{2,000} & 0,759 & 0,819 & 0,726 & 0,411 & 0,278 & 0,367 \\
\textit{no\_lateral\_lenses} & 1,533 & \textbf{0,793} & \textbf{0,845} & \textbf{0,728} & 0,561 & 0,423 & 0,467 \\
\textit{full\_system} & 1,767 & 0,770 & 0,827 & 0,719 & 0,444 & 0,381 & 0,467 \\
\bottomrule
\end{tabular}%
}
\caption*{\footnotesize Fuente/procedencia: elaboración propia. La fluidez representa el número medio de hipótesis extraídas por ejecución exitosa y evaluable; la tasa SVHN resume hipótesis estructuralmente válidas y de alta novedad; VCN corresponde a novedad combinatoria válida. En negrita se marca el mejor valor de cada columna.}
\end{table*}

\begin{table*}[!tp]
\centering
\scriptsize
\caption{Tasa de finalización y fallos técnicos por condición.}
\label{tab:appendix-completion-es}
\resizebox{\textwidth}{!}{%
\begin{tabular}{lccccccc}
\toprule
\textbf{Condición} &
\textbf{Esperadas} &
\textbf{Completadas} &
\textbf{Fallidas} &
\textbf{Tasa complet.} &
\textbf{Problemas API} &
\textbf{Límite API} &
\textbf{Otros técnicos} \\
\midrule
\textit{direct\_llm} & 30 & 30 & 0 & \textbf{1,000} & 0,000 & 0,000 & 0,000 \\
\textit{self\_reflection} & 30 & 30 & 0 & \textbf{1,000} & 0,000 & 0,000 & 0,000 \\
\textit{no\_agent4} & 30 & 30 & 0 & \textbf{1,000} & 0,000 & 0,000 & 0,000 \\
\textit{no\_search} & 30 & 30 & 0 & \textbf{1,000} & 0,000 & 0,000 & 0,000 \\
\textit{no\_lateral\_lenses} & 30 & 30 & 0 & \textbf{1,000} & 0,000 & 0,000 & 0,000 \\
\textit{full\_system} & 30 & 30 & 0 & \textbf{1,000} & 0,000 & 0,000 & 0,000 \\
\bottomrule
\end{tabular}%
}
\caption*{\footnotesize Fuente/procedencia: elaboración propia. La fila técnica \textit{join\_failed} del manifiesto no forma parte del diseño experimental y se excluye de esta tabla.}
\end{table*}

\begin{table*}[!tp]
\centering
\scriptsize
\caption{Diferencias pareadas del sistema completo frente a cada referencia.}
\label{tab:appendix-paired-deltas-es}
\resizebox{\textwidth}{!}{%
\begin{tabular}{llccccc}
\toprule
\textbf{Comparador} &
\textbf{Métrica} &
\textbf{Bloques} &
\(\boldsymbol{\Delta}\) \textbf{medio} &
\textbf{Tasa victoria} &
\textbf{Cohen \(d_z\)} &
\textbf{IC 95\%} \\
\midrule
\textit{direct\_llm} & Factibilidad & 3 & 0,139 & 1,000 & 1,981 & [0,075; 0,214] \\
\textit{direct\_llm} & Originalidad & 3 & 0,263 & 1,000 & 7,896 & [0,235; 0,300] \\
\textit{direct\_llm} & Rareza comb. & 3 & 0,357 & 1,000 & 1,265 & [0,068; 0,631] \\
\textit{direct\_llm} & Diversidad & 2 & 0,006 & 0,500 & 0,064 & [-0,056; 0,067] \\
\textit{direct\_llm} & Colapso & 2 & -0,031 & 0,500 & -0,193 & [-0,146; 0,083] \\
\textit{direct\_llm} & Creatividad & 3 & 0,153 & 1,000 & 3,774 & [0,108; 0,187] \\
\textit{self\_reflection} & Factibilidad & 3 & 0,021 & 0,667 & 0,346 & [-0,044; 0,078] \\
\textit{self\_reflection} & Originalidad & 3 & 0,031 & 1,000 & 2,033 & [0,013; 0,041] \\
\textit{self\_reflection} & Rareza comb. & 3 & 0,019 & 0,667 & 0,231 & [-0,073; 0,078] \\
\textit{self\_reflection} & Diversidad & 3 & -0,005 & 0,667 & -0,113 & [-0,060; 0,025] \\
\textit{self\_reflection} & Colapso & 3 & -0,021 & 0,333 & -0,375 & [-0,071; 0,038] \\
\textit{self\_reflection} & Creatividad & 3 & 0,006 & 0,667 & 0,283 & [-0,013; 0,029] \\
\textit{no\_agent4} & Factibilidad & 3 & 0,011 & 0,667 & 0,296 & [-0,028; 0,047] \\
\textit{no\_agent4} & Originalidad & 3 & 0,045 & 1,000 & 0,633 & [0,002; 0,128] \\
\textit{no\_agent4} & Rareza comb. & 3 & 0,091 & 0,667 & 0,605 & [-0,017; 0,264] \\
\textit{no\_agent4} & Diversidad & 3 & 0,065 & 1,000 & 0,777 & [0,000; 0,159] \\
\textit{no\_agent4} & Colapso & 3 & -0,046 & 0,000 & -2,639 & [-0,062; -0,028] \\
\textit{no\_agent4} & Creatividad & 3 & 0,023 & 1,000 & 0,784 & [0,003; 0,058] \\
\textit{no\_search} & Factibilidad & 3 & -0,011 & 0,333 & -0,158 & [-0,086; 0,053] \\
\textit{no\_search} & Originalidad & 3 & 0,001 & 0,667 & 0,036 & [-0,037; 0,028] \\
\textit{no\_search} & Rareza comb. & 3 & 0,103 & 1,000 & 1,014 & [0,037; 0,220] \\
\textit{no\_search} & Diversidad & 3 & -0,031 & 0,333 & -0,291 & [-0,123; 0,087] \\
\textit{no\_search} & Colapso & 3 & -0,002 & 0,667 & -0,042 & [-0,052; 0,026] \\
\textit{no\_search} & Creatividad & 3 & -0,007 & 0,333 & -0,189 & [-0,036; 0,033] \\
\textit{no\_lateral\_lenses} & Factibilidad & 3 & -0,049 & 0,000 & ---\textsuperscript{*} & [-0,050; -0,047] \\
\textit{no\_lateral\_lenses} & Originalidad & 3 & 0,006 & 0,333 & 0,090 & [-0,040; 0,079] \\
\textit{no\_lateral\_lenses} & Rareza comb. & 3 & 0,178 & 1,000 & 3,924 & [0,147; 0,230] \\
\textit{no\_lateral\_lenses} & Diversidad & 3 & -0,027 & 0,333 & -0,428 & [-0,085; 0,042] \\
\textit{no\_lateral\_lenses} & Colapso & 3 & -0,015 & 0,333 & -0,283 & [-0,062; 0,042] \\
\textit{no\_lateral\_lenses} & Creatividad & 3 & -0,009 & 0,333 & -0,223 & [-0,039; 0,034] \\
\bottomrule
\end{tabular}%
}
\caption*{\footnotesize Fuente/procedencia: elaboración propia. Se define \(\Delta = m_{\text{sistema}} - m_{\text{referencia}}\) dentro del mismo bloque problema--semilla. La tasa de victoria es siempre la fracción de bloques con \(\Delta > 0\); en la tasa de colapso, por tanto, una tasa de victoria baja y \(\Delta\) negativo favorecen al sistema completo, porque indican menor colapso semántico. Estos valores no deben leerse como evidencia inferencial fuerte: con solo tres bloques efectivos, y dos en alguna métrica, \(d_z\), la tasa de victoria y el IC 95\% son descriptores exploratorios sensibles a pequeñas variaciones y a varianzas casi nulas. \textsuperscript{*}\(d_z\) se omite cuando la varianza entre bloques es casi nula y el valor resultante carece de interpretación útil.}
\end{table*}

\begin{table*}[!tp]
\centering
\scriptsize
\caption{Ranking medio por condición en métricas seleccionadas.}
\label{tab:appendix-average-ranks-es}
\resizebox{\textwidth}{!}{%
\begin{tabular}{lccccccc}
\toprule
\textbf{Condición} &
\textbf{Validez estr.} &
\textbf{Factibilidad} &
\textbf{Originalidad} &
\textbf{Rareza comb.} &
\textbf{Diversidad} &
\textbf{Colapso} &
\textbf{Creatividad} \\
\midrule
\textit{direct\_llm} & 4,667 & 6,000 & 6,000 & 5,667 & 4,000 & 3,500 & 6,000 \\
\textit{self\_reflection} & \textbf{1,667} & \textbf{2,333} & 4,000 & 2,000 & \textbf{3,000} & 3,667 & 3,333 \\
\textit{no\_agent4} & 3,000 & 3,333 & 4,000 & 3,000 & 4,333 & 4,667 & 4,333 \\
\textit{no\_search} & 5,000 & 3,667 & 3,000 & 4,000 & \textbf{3,000} & \textbf{2,333} & 2,667 \\
\textit{no\_lateral\_lenses} & 2,667 & \textbf{2,333} & \textbf{2,000} & 4,667 & \textbf{3,000} & 3,333 & \textbf{2,000} \\
\textit{full\_system} & 4,000 & 3,333 & \textbf{2,000} & \textbf{1,667} & \textbf{3,000} & 2,667 & 2,667 \\
\bottomrule
\end{tabular}%
}
\caption*{\footnotesize Fuente/procedencia: elaboración propia. El ranking se calcula dentro de bloques problema--semilla; valores menores indican mejor posición relativa. En \textit{Colapso}, también se interpreta mejor un ranking menor porque corresponde a menor \textit{CollapseRate}; en esa métrica \textit{direct\_llm} solo participa en los 2 bloques donde produjo al menos dos hipótesis. En negrita se marca el mejor ranking medio de cada métrica.}
\end{table*}

\begin{table*}[!tp]
\centering
\scriptsize
\caption{Métricas principales desagregadas por problema experimental.}
\label{tab:appendix-per-problem-es}
\resizebox{\textwidth}{!}{%
\begin{tabular}{llcccccc}
\toprule
\textbf{Problema} &
\textbf{Condición} &
\textbf{Validez estructural} &
\textbf{Tasa factible} &
\textbf{Originalidad} &
\textbf{Rareza comb.} &
\textbf{Tasa SVHN} &
\textbf{Calidad} \\
\midrule
Desalinización & \textit{direct\_llm} & 0,367 & 0,617 & 0,350 & 0,651 & 0,067 & 0,725 \\
Desalinización & \textit{self\_reflection} & \textbf{0,750} & 0,925 & 0,601 & 0,869 & \textbf{0,750} & 0,794 \\
Desalinización & \textit{no\_agent4} & 0,483 & 0,975 & 0,623 & 0,866 & 0,433 & 0,780 \\
Desalinización & \textit{no\_search} & 0,483 & 0,904 & 0,632 & 0,743 & 0,483 & 0,756 \\
Desalinización & \textit{no\_lateral\_lenses} & 0,700 & \textbf{1,000} & \textbf{0,658} & 0,678 & 0,700 & \textbf{0,812} \\
Desalinización & \textit{full\_system} & 0,567 & 0,933 & 0,627 & \textbf{0,871} & 0,567 & 0,784 \\
\midrule
Parlamento & \textit{direct\_llm} & 0,500 & 0,800 & 0,468 & 0,171 & 0,300 & 0,758 \\
Parlamento & \textit{self\_reflection} & 0,467 & \textbf{1,000} & 0,663 & 0,751 & 0,467 & \textbf{0,791} \\
Parlamento & \textit{no\_agent4} & \textbf{0,533} & 0,950 & 0,575 & 0,538 & \textbf{0,533} & 0,773 \\
Parlamento & \textit{no\_search} & 0,267 & \textbf{1,000} & 0,690 & 0,750 & 0,267 & 0,765 \\
Parlamento & \textit{no\_lateral\_lenses} & 0,283 & \textbf{1,000} & 0,625 & 0,656 & 0,283 & 0,756 \\
Parlamento & \textit{full\_system} & 0,200 & 0,967 & \textbf{0,704} & \textbf{0,802} & 0,200 & 0,741 \\
\bottomrule
\end{tabular}%
}
\caption*{\footnotesize Fuente/procedencia: elaboración propia. Medias por ejecución agregadas por problema y condición; desalinización combina los bloques de semillas 42 y 43, parlamento corresponde a la semilla 42. En negrita se marca el mejor valor de cada métrica dentro de cada problema. La validez estructural relativamente alta de \textit{direct\_llm} en parlamento convive con la rareza combinatoria más baja de todo el experimento (0,171): produce propuestas convencionales y por ello fáciles de validar.}
\end{table*}

\subsection*{Ejemplos de salidas del sistema}
La Tabla~\ref{tab:appendix-qualitative-examples-es} amplía la trazabilidad de la Tabla~\ref{tab:qualitative-examples-es} y recoge ejemplos adicionales de las condiciones con ablaciones. Estos casos no deben leerse como una jerarquía absoluta, sino como evidencia de modos de comportamiento distintos: algunas ablaciones producen soluciones muy sólidas, mientras que otras revelan con claridad el papel del anclaje empírico o del filtrado final.

\begin{table*}[!tp]
\centering
\scriptsize
\caption{Procedencia y ejemplos cualitativos adicionales por condición experimental.}
\label{tab:appendix-qualitative-examples-es}
\resizebox{\textwidth}{!}{%
\begin{tabular}{p{1.7cm}p{2.2cm}p{3.8cm}p{6.1cm}p{4.0cm}}
\toprule
\textbf{Uso} &
\textbf{Condición} &
\textbf{Hipótesis} &
\textbf{Lectura cualitativa} &
\textbf{Procedencia y métricas} \\
\midrule
Principal &
\textit{direct\_llm} &
\textbf{Bombeo electromagnético de agua de mar en estado sólido} &
Idea llamativa pero mal orientada: sustituye el bombeo mecánico por MHD sin resolver directamente la separación iónica. Sirve como ejemplo de especulación atractiva con baja constricción física. &
Ejecución 0065, id 3. Desalinización, semilla 43, repetición 05. \(F=1\), \(Q=0{,}528\). \\
\addlinespace
Principal &
\textit{self\_reflection} &
\textbf{Desalinización por congelación acústica resonante} &
La autorreflexión mejora mucho la forma final y alcanza alta puntuación, pero conserva una dependencia fuerte de PEF subnanosegundo y nucleación acústica precisa. &
Ejecución 0073, id 2. Desalinización, semilla 43, repetición 03. \(F=3\), \(Q=1{,}000\). \\
\addlinespace
Principal &
\textit{full\_system} &
\textbf{Desalinización crioflotante con clatratos de ciclopentano} &
La crítica externa desplaza el diseño hacia un mecanismo de separación más pasivo: clatratos de ciclopentano, flotabilidad y gestión explícita del calor latente. &
Ejecución 0117, id 1. Desalinización, semilla 43, repetición 07. \(F=3\), \(Q=0{,}917\). \\
\addlinespace
Principal &
\textit{direct\_llm} &
\textbf{Voto delegado meritocrático con pruebas de conocimiento cero} &
Introduce un oráculo de mérito que altera el peso del voto parlamentario. La propuesta es sofisticada en superficie, pero políticamente frágil y estructuralmente inválida. &
Ejecución 0129, id 1. Parlamento, semilla 42, repetición 09. \(F=1\), \(Q=0{,}694\). \\
\addlinespace
Principal &
\textit{self\_reflection} &
\textbf{Torneos aleatorizados de predicción de evidencia} &
Ejemplo fuerte de autorreflexión: propone un mecanismo asesor implementable para mejorar la evaluación de evidencia, aunque no ataca directamente la coerción partidista. &
Ejecución 0137, id 3. Parlamento, semilla 42, repetición 07. \(F=3\), \(Q=1{,}000\). \\
\addlinespace
Principal &
\textit{full\_system} &
\textbf{Depósito epistémico electoral con resolución adversarial} &
La salida final evita ponderar votos o usar secreto parlamentario y desplaza el incentivo al coste interno de partido, manteniendo igualdad formal de los diputados. &
Ejecución 0171, id 1. Parlamento, semilla 42, repetición 01. \(F=2\), \(Q=0{,}889\). \\
\midrule
Apéndice &
\textit{no\_search} &
\textbf{Clasificador balístico supersónico de iones por MHD} &
Sin búsqueda externa reaparece una solución MHD espectacular: campos \(>10\,\mathrm{T}\), presión oceánica y deflexión Lorentz. Es creativa, pero muy vulnerable a mezcla turbulenta y escala física. &
Ejecución 0098, id 1. Desalinización, semilla 43, repetición 08. \(F=2\), \(Q=0{,}806\). \\
\addlinespace
Apéndice &
\textit{no\_agent4} &
\textbf{Polarización microfluídica de concentración iónica} &
La ausencia del último agente no impide producir una hipótesis sólida: reduce MHD macroscópico a electrocinética microfluídica, una ruta físicamente más defendible. &
Ejecución 0022, id 3. Desalinización, semilla 42, repetición 02. \(F=3\), \(Q=0{,}917\). \\
\addlinespace
Apéndice &
\textit{no\_lateral\_lenses} &
\textbf{Esponja electroquímica de iones basada en análogos de azul de Prusia} &
Al eliminar lentes laterales aparece una solución menos exótica y más cercana a tecnologías existentes de intercalación electroquímica. Gana plausibilidad, pero pierde parte del salto combinatorio. &
Ejecución 0042, id 1. Desalinización, semilla 42, repetición 02. \(F=3\), \(Q=0{,}917\). \\
\addlinespace
Apéndice &
\textit{no\_search} &
\textbf{Mercado cuadrático de tiempo para controlar la agenda} &
Propone una economía procedimental donde el coste de acaparar tiempo parlamentario crece cuadráticamente. Es elegante, aunque depende de una formalización institucional no contrastada externamente. &
Ejecución 0155, id 1. Parlamento, semilla 42, repetición 05. \(F=3\), \(Q=0{,}917\). \\
\addlinespace
Apéndice &
\textit{no\_agent4} &
\textbf{Activador del artículo 85.1 con conocimiento cero} &
Ejemplo especialmente valioso: usa un anclaje jurídico local, el umbral de 70 firmas del artículo 85.1, y aplica criptografía solo para proteger coordinación interna. &
Ejecución 0150, id 1. Parlamento, semilla 42, repetición 10. \(F=3\), \(Q=0{,}917\). \\
\addlinespace
Apéndice &
\textit{no\_lateral\_lenses} &
\textbf{Dividendo procedimental ponderado por consenso} &
Solución sobria y probablemente implementable: recompensa el consenso multipartidista con vía rápida procedimental. Muestra que retirar lateralidad puede aumentar claridad institucional. &
Ejecución 0167, id 2. Parlamento, semilla 42, repetición 07. \(F=3\), \(Q=0{,}917\). \\
\bottomrule
\end{tabular}%
}
\caption*{\footnotesize Fuente/procedencia: elaboración propia a partir de registros experimentales y salidas JSON reales. \(F\) corresponde a la escala de factibilidad discreta y \(Q\) a la puntuación de calidad.}
\end{table*}

\printbibliography[title={References / Bibliografía y webgrafía}]

\end{document}